\documentclass[lettersize,journal]{IEEEtran}
\ifx\pdfoutput\undefined\else\pdfoutput=1\fi
\usepackage{amsmath,amsfonts}
\usepackage{algorithmic}
\usepackage{algorithm}
\usepackage{array}
\usepackage[caption=false,font=normalsize,labelfont=sf,textfont=sf]{subfig}
\usepackage{textcomp}
\usepackage{url}
\usepackage{verbatim}
\usepackage{graphicx}
\usepackage[hidelinks]{hyperref}
\usepackage{tabularray}
\usepackage{float}
\usepackage[T1]{fontenc}
\usepackage[utf8]{inputenc}
\usepackage{ulem}
\usepackage{caption}
\usepackage{soul}
\usepackage{xcolor}
\usepackage{tabularx}
\usepackage{makecell}
\usepackage{booktabs}
\usepackage{multirow}
\usepackage{colortbl}
\usepackage{siunitx}
\hypersetup{
	colorlinks=false,
	pdfborder={0 0 0},
	urlbordercolor={1 1 1},
	linkbordercolor={1 1 1},
	citebordercolor={1 1 1}
}
\UseTblrLibrary{booktabs}
\definecolor{lightorange}{RGB}{255, 220, 180}
\definecolor{lightblue}{RGB}{180, 215, 255}

\newcommand{\hlorange}[1]{\colorbox{lightorange}{#1}}
\newcommand{\hlblue}[1]{\colorbox{lightblue}{#1}}

\begin{document}
	
	\title{Zero-shot Tweet-Level Stance Detection Enhanced by External Knowledge and Reflective Chain-of-Thought Reasoning}
	
	\author{Yiju Huang,~Wenxian Wang,~Lijun Zhou,~Rui Tang,~Xiao Lan,~Tao Zhang,~\textit{and}~Haizhou Wang\textsuperscript{*}
		\thanks{Yiju Huang, Wenxian Wang, Lijun Zhou, Rui Tang, Xiao Lan, and Haizhou Wang are with the School of Cyber Science and Engineering, Sichuan University, Chengdu, China.}%
		\thanks{Tao Zhang is with the School of Cyberspace Science and Technology, Beijing Jiaotong University, Beijing, China.}%
		\thanks{Corresponding author: Haizhou Wang (e-mail: whzh.nc@scu.edu.cn).}%
		\thanks{Manuscript received April 19, 2021; revised August 16, 2021.}}
	
	\markboth{}{Huang \MakeLowercase{\textit{et al.}}: Zero-shot Tweet-Level Stance Detection Enhanced by External Knowledge and Reflective CoT Reasoning}
	
	
	\maketitle
	
	\begin{abstract}
		Zero-shot tweet-level stance detection confronts two primary challenges: (1) mitigating the context sparsity inherent in short texts, and (2) establishing the relevance between implicit targets and textual content. While existing methods primarily focus on incorporating external knowledge, they neglect the intrinsic semantic cues embedded within key intra-textual entities. Furthermore, current models exhibit limited capability in determining the relevance of unseen targets to the given text, thereby struggling to differentiate between “neutral” and “irrelevant” stance labels.
		
		To address these issues, we first construct a four-class, multi-topic Japanese tweet dataset. To our knowledge, this is the first Japanese tweet-level dataset for stance detection. We then propose KIRP, a zero-shot stance detection framework. It integrates external knowledge with entity reorganization for data augmentation and employs prompt chaining for reasoning. Specifically, the framework incorporates knowledge graphs to supplement and reorganize key textual entities, while reflective Chain-of-Thought (CoT) reasoning extracts and validates implicit targets. To better distinguish “neutral” from “irrelevant” labels, we adopt stance-aware contrastive learning to capture discriminative features and design a three-layer iterative prototype network for fine-grained classification. Experimental results on SemEval-2016, WT-WT, and KIRP-D show that KIRP achieves state-of-the-art performance. KIRP obtains F1 scores of 84.05\% (three-class) on SemEval-2016, and 84.99\% and 79.18\% (four-class) on WT-WT and KIRP-D, respectively.
	\end{abstract}
	
	
	\section{Introduction}
	\IEEEPARstart{S}{tance} detection identifies the attitude expressed in a text toward a specific target, typically categorized as favor, against, neutral, or irrelevant. This task is essential for real-time applications such as market analysis and public opinion monitoring. Crucially, stance is highly dependent on the target. In social media, a shift in the target often changes the expressed stance. For instance, regarding the target “sewage discharge” the tweet “Discharging sewage into the sea is terrible; I no longer dare to eat seafood” expresses an “against” stance. However, if the target changes to “environmental protection”, the same tweet conveys a “favor” stance.
	
	Zero-Shot Stance Detection (ZSSD) is an emerging research direction that performs well under data-scarce conditions. Its core principle is to enable knowledge transfer through shared representations, often learned via meta-learning or transfer learning. This allows models to apply related knowledge structures to new domains. However, tweet-level ZSSD is more challenging than conventional settings. Tweets often lack contextual information and contain implicit targets, requiring more effective and precise integration of external knowledge.
	
	Tweets often suffer from limited or missing entity information. However, entity augmentation and reorganization can generate semantically richer representations. Transforming valid entities also produces additional target-related information, strengthening the association between the text and the intended target. TABLE I illustrates these two processes. In the entity augmentation example, the blue-highlighted segment identifies an entity to be augmented. By incorporating related entities from a knowledge graph, we can make implicit targets explicit. For instance, a tweet might use the pronoun “it” without specifying the object of criticism. If this tweet appears under the topic “nuclear energy development”, our framework augments “it” with “nuclear energy.” This identifies the user's “against” stance toward the target, which would otherwise be difficult to interpret without prior knowledge.
	
		\begin{table}[h]
			\centering
			\caption{Examples of entity augmentation and reorganization, where words highlighted in blue represent the completed or replaced parts, and words highlighted in orange indicate the parts that can be completed.}
			\renewcommand{\arraystretch}{2.0}
			\begin{tabularx}{\columnwidth}{|
					>{\centering\arraybackslash}m{2.2cm}|
					>{\raggedright\arraybackslash}X|
				}
				\hline
				\textbf{Type} & \textbf{Text} \\
				\hline
				
				\makecell[c]{Original\\Text} &
				It is a waste of electricity and water. Also golf courses waste water \\
				\hline
				
				\makecell[c]{Entity\\Augmentation} &
				\hlorange{Nuclear} is a waste of electricity and water. Also golf courses waste water \\
				
				\hline
				
				\makecell[c]{Entity\\Reorganization} &
				It is a waste of electricity and water. Also \hlblue{football fields} waste water \\
				
				\hline
			\end{tabularx}
		\end{table}
	
	In the entity reorganization example, the orange-highlighted segment identifies valid entities for reorganization. By matching equivalent entities in a knowledge graph, we generate additional target information. This expands the scope of LLM prompt chains and strengthens the alignment between the text and the intended target. For instance, while the original sentence targets “building sports facilities”, modifying specific entities can shift the stance. By replacing “golf course” with its equivalent “football field” via knowledge graph-based substitution, the model can evaluate different stance values under the same target.
	
	Early studies utilized rule-based and feature engineering approaches for stance detection~\cite{sobhani2016detecting}~\cite{augenstein2016stance}~\cite{sobhani2017dataset}. However, these methods rely on superficial lexical or sentiment features, failing to capture deep semantic relationships. Subsequent efforts based on weakly supervised learning~\cite{dias2016heuristics}~\cite{wei2019topic}~\cite{kumar2021weakly} reduced the need for manual annotation. Nevertheless, these models often suffer from label noise and limited data, resulting in inaccurate feature extraction and poor generalization.
	
	Some early studies employed hybrid architectures combining CNN and LSTM for stance detection~\cite{kumar2024negative}. However, these methods often overlook target variations during training, making detection results overly dependent on textual content. Consequently, these models struggle to identify implicit targets and perform poorly on unseen targets.
	
	Recent research has shifted from large-scale supervised training toward feature sharing and transfer learning for data-scarce environments~\cite{hardalov2022few}~\cite{aiyappa2024benchmarking}~\cite{jiang2022few}~\cite{liang2022zero}. To mitigate information loss, current methods utilize external knowledge injection, adversarial learning, and LLMs for rich semantic representations. Some approaches employ multi-dataset learning and pattern-based data transformations ~\cite{hardalov2022few} to provide additional context. Others incorporate instruction-tuned LLMs~\cite{aiyappa2024benchmarking} and target-aware prompt distillation~\cite{jiang2022few} for fine-tuning. Furthermore, prompt learning and contrastive learning~\cite{liang2022zero} are used to capture correlations between target-invariant features and stance labels, facilitating effective knowledge transfer.
	
	However, existing studies still exhibit weak alignment between targets and textual content, making it difficult to distinguish “neutral” from “irrelevant” stances. Furthermore, the reasoning capabilities of large models remain underutilized, limiting performance on short, tweet-level texts.
	
	In summary, zero-shot tweet-level stance detection faces three primary challenges: (1) supplementing external knowledge within sparse short texts; (2) extracting valid implicit targets; and (3) assessing relevance between unseen targets and text to differentiate “neutral” from “irrelevant” labels. Our proposed framework, KIRP, provides a comprehensive solution to these challenges.
	
	Specifically, our contributions are as follows:
	\begin{itemize}
		\item \textbf{We construct KIRP-D, a four-class, multi-topic dataset for data-scarce environments.} To our knowledge, this is the first Japanese dataset specifically designed for zero-shot stance detection. KIRP-D supports cross-target feature transfer and implicit target discovery, bridging a critical gap in Japanese stance detection corpora and providing diverse, up-to-date topical data for future research.
		\item \textbf{We introduce an entity reorganization-based data augmentation method within a LLM-based reflective chain-of-thought (CoT) reasoning framework.} By aligning textual entities with equivalent knowledge from external sources, our method generates diverse target representations. This approach effectively uncovers implicit targets and expands the LLM's reasoning scope, yielding more precise stance predictions and robust supporting evidence.
		\item \textbf{We propose a KIRP framework that leverages reflective CoT prompting to enhance reasoning and employs a prototypical network for classification.} By combining knowledge injection with reorganization-based augmentation, KIRP generates rich semantic representations via reflective CoT reasoning. Furthermore, our stance-aware contrastive learning optimizes the representation space, followed by a three-layer prototype network for iterative four-class classification.
		\item \textbf{We extensively validate KIRP and KIRP-D, demonstrating state-of-the-art (SOTA) performance.} KIRP achieves an F1 score of 84.05\% in three-class tasks and scores of 84.99\% and 79.18\% in four-class tasks on WT-WT and KIRP-D, respectively. Experimental results further confirm the generalizability of KIRP-D across other cross-target stance detection settings.
	\end{itemize}
	
	\section{RELATED WORK}
	Zero-shot stance detection identifies text stances toward unseen targets. Recently, research has converged on three primary technical paths: (1) injecting external knowledge to compensate for missing target information; (2) employing contrastive learning to extract transferable features; and (3) leveraging LLM reasoning for zero-shot prediction without task-specific training. The following sections review these approaches and the relevant literature.

\subsection{Methods Based on External Knowledge Injection}
Knowledge injection provides prior information to enhance features and reduce dependency on labeled data. This approach improves generalization and semantic understanding, making it a promising solution for data-scarce environments. Existing studies have integrated knowledge graphs into pre-trained models, significantly improving stance detection for implicit targets~\cite{lin2024kpatch,zhang2024commonsense}. Furthermore, researchers have explored multimodal information. For instance, Liang et al.~\cite{liang2024multi} fused text and image data to identify user stances, while Khiabani et al.~\cite{khiabani2023few} proposed a multimodal embedding model combining textual and network features for cross-target detection.
	
	Moreover, several studies emphasize the role of sentiment information in stance detection, integrating it to bolster model performance. Zhou et al. ~\cite{zhou2021topicbert} introduced TopicBERT, a pre-trained model designed to enhance BERT's effectiveness in sentiment-related tasks. Garg et al. ~\cite{garg2024stanceformer} proposed Stanceformer, a sentiment-enhanced Transformer architecture that addresses the underutilization of target information in conventional Transformers. Furthermore, Sun et al. ~\cite{sun2022stance} leveraged adversarial networks and attention mechanisms to joint-learn topic and sentiment features, thereby improving stance detection accuracy.
	
	Beyond conventional knowledge, recent research explores the impact of ancillary features—such as temporal data and user demographics—on stance detection. Li et al. ~\cite{li2024pro} integrated gender bias factors as external knowledge to refine stance analysis. Zhang et al. ~\cite{zhang2024enhancing} incorporated moral dimensions as quantifiable metrics, enhancing the model’s ability to predict user attitudes. Additionally, Mu et al. ~\cite{mu2024examining} introduced temporal factors into COVID-19 vaccine stance detection, demonstrating the necessity of considering time-series dynamics in social media data analysis.
	
	Knowledge injection is vital for stance detection in data-scarce environments, enabling models to leverage prior knowledge for reasoning. However, acquiring high-quality knowledge is challenging. Issues such as limited coverage, obsolete information, and inaccuracies can introduce noise, undermining detection accuracy. Although progress has been made in formalizing knowledge, existing methods still struggle to represent the complex semantics and structures of diverse knowledge sources. Specifically, current representations often fail to capture ambiguity or context-dependent meanings, leading to ineffective injection and diminished model performance.
	
	\subsection{Methods Based on Contrastive Learning}Contrastive learning facilitates the extraction of richer, more representative features by distinguishing between similar and dissimilar samples. In stance detection, this approach captures fine-grained feature differences at semantic and syntactic levels, deepening the model's understanding of the text-target relationship. Even in data-scarce environments, contrastive learning bolsters the model’s ability to extract discriminative, stance-related features, making it a robust tool for zero-shot scenarios.
	
	Several studies have integrated contrastive learning into stance detection. These approaches typically employ contrastive modules to aggregate samples with the same polarity and separate those with different polarities, thereby maximizing the discriminative representation space while minimizing intra-class distance. Yao et al. ~\cite{yao2024enhancing} combined contrastive learning with prompt learning, dynamically adjusting prompt influence to enhance zero-shot detection. Wang et al. ~\cite{wang2024meta} jointly optimized classification and contrastive losses to strengthen the model’s ability to differentiate stances. Furthermore, Zhao et al. ~\cite{zhao2024msfr} proposed a hierarchical contrastive learning framework to capture multi-level semantic features, facilitating knowledge transfer from known to unseen targets.
	
	While most studies use stance polarity as the primary contrastive criterion, multi-perspective networks are emerging to capture latent features. However, these methods may introduce noise or increase computational overhead. Jiang et al. ~\cite{jiang2023zero} proposed a multi-view contrastive learning framework that leverages both labeled and unlabeled data to enhance unseen target representations, extracting more discriminative features for zero-shot detection. Similarly, Zhao et al. ~\cite{zhao2023feature} utilized multi-view contrastive learning to capture target-invariant syntactic patterns. They combined these patterns with BERT-encoded masked sentences to extract syntactic features, further improving stance prediction.
	
	The selection of negative samples is critical to the efficacy of contrastive learning. However, data-scarce environments often lack a sufficient pool of suitable negatives. Poorly selected samples—either too similar to or too distinct from positive instances—can lead to suboptimal feature learning, undermining stance detection performance. To address this, our method generates a diverse set of negative samples through entity reorganization, effectively bolstering contrastive learning performance by providing high-quality, target-oriented variations.
	
	\subsection{Methods Based on Large Language Model}Large Language Models (LLMs) have emerged as a dominant research paradigm due to their extensive linguistic knowledge and robust semantic understanding, acquired through pre-training on massive corpora. By employing prompt chaining, researchers can guide LLMs to leverage this prior knowledge for zero-shot stance detection. Even in data-scarce environments, LLMs effectively infer stances by analyzing textual context against their internal knowledge base. Furthermore, prompt chains offer the flexibility to be tailored for specific scenarios, allowing LLMs to adapt seamlessly to diverse and complex stance detection tasks.
	
	Zhao et al. ~\cite{zhao2024zerostance} utilized ChatGPT to generate CHATStance, a synthetic open-domain dataset for training models with cross-target generalization. Lan et al. ~\cite{lan2024stance} proposed COLA, a three-stage framework employing collaborative LLM agents to identify latent stance cues. For reasoning-based analysis, Ding et al. ~\cite{ding2024cross} introduced TsCoT, a two-stage chain-of-thought (CoT) method that generates multi-perspective explanations. In a subsequent study, Ding et al. ~\cite{ding2025adversarial} integrated Wikipedia-based background knowledge with CoT reasoning to ensure the timeliness and relevance of target information. Despite these advancements, existing studies still underutilize the deep reasoning potential of LLMs and often struggle to adapt them effectively to specialized domain tasks.
	
	In this work, we employ an improved counterfactual prompt chain to generate reasoning evidence across four perspectives: support, against, neutral, and irrelevant. This process provides rich contextual semantics and reasoning cues for short texts. The resulting textual rationales are then fed into a pre-trained model to learn deep, stance-related textual representations.
	
	\section{DATASET}
	\subsection{Dataset Overview}
	Stance detection is a mature NLP task with a decade-long history of diverse dataset development. Most datasets consist of large-scale English corpora, which are well-suited for traditional data-driven machine learning and deep learning approaches. While these methods perform strongly on abundant data, recent research highlights the catastrophic performance degradation of traditional models in data-scarce environments. This has shifted focus toward few-shot stance detection, leading to the creation of high-quality datasets like VAST~\cite{allaway2020zero}.
	
	These few-shot datasets typically feature a wide range of topics with minimal samples per topic, mimicking the data sparsity of real-world, topic-specific scenarios. Consequently, models must demonstrate robust transfer and generalization capabilities to achieve competitive results. Key characteristics of several mainstream stance detection datasets are summarized in TABLE II.
	\begin{table*}[!t]
		\centering
		\caption{Main Stance Detection Datasets}
		\label{tab:datasets}
		
		\begin{tblr}{
				width=\textwidth,
				colspec={l c l l c},
				row{1} = {font=\bfseries},
				row{13} = {font=\bfseries},
				hline{1,13,14} = {-}{},
				hline{2} = {-}{}
			}
			
			Dataset & Year & Content & Language / Size & Classification \\
			
			SemEval-2016 T6~\cite{mohammad2016semeval} & 2016 & Specific topics & English / $\sim$4,800 & Three-class \\
			
			Multi-target~\cite{sobhani2017dataset} & 2017 & U.S. election & English / 4,455 & Three-class \\
			
			WT-WT~\cite{conforti2020will} & 2020 & Financial domain & English / 51,284 tweets & Four-class \\
			
			VAST~\cite{allaway2020zero} & 2020 & \textit{New York Times} debates & English / 18,515 comments & Three-class \\
			
			P-Stance~\cite{li2021p} & 2021 & U.S. election & English / 21,574 tweets & Three-class \\
			
			CoVaxNet~\cite{jiang2022covaxnet} & 2023 & COVID-19 vaccines & English / 1.8M tweets & Unlabeled \\
			
			ISD~\cite{huang2023knowledge} & 2023 & U.S. election & English / 6,027 annotations & Three-class \\
			
			ProCons~\cite{wang2023quantifying} & 2023 & Weibo content & Chinese / 32,667 posts & Three-class \\
			
			C-Stance~\cite{zhao2023c} & 2023 & Weibo content & Chinese / 48,126 annotations & Three-class \\
			
			MMVAX-Stance~\cite{weinzierl2023identification} & 2023 & COVID-19 vaccines & English / 11,300 annotations & Three-class \\
			
			MT-CSD~\cite{niu2024challenge} & 2024 & Reddit discussions & English / 15,876 annotations & Three-class \\
			
			ours & 2026 & X Tweets & Japanese / 43,909 annotations & Four-class \\
			
		\end{tblr}
		
	\end{table*}
	
	Despite the proliferation of stance detection datasets, existing benchmarks still exhibit several notable limitations: 
	\begin{enumerate}
	\item \textbf{Language imbalance}: current resources are heavily skewed toward English and Chinese, while other languages such as Japanese remain largely underrepresented, creating a significant gap in multilingual stance detection research. 
	
	\item \textbf{Limited annotation granularity}: with the exception of WT-WT (2020), most existing datasets adopt a three-class annotation scheme (favor, against, neutral), overlooking the practically important “irrelevant” category where a text bears no relation to the given target. 
	
	\item \textbf{Topic diversity}: many benchmarks concentrate on a narrow set of domains, which may restrict the cross-domain generalizability of models trained on such data.
	\end{enumerate}
	
	
	\begin{figure}[!t]
		\centering
		\includegraphics[width=\linewidth]{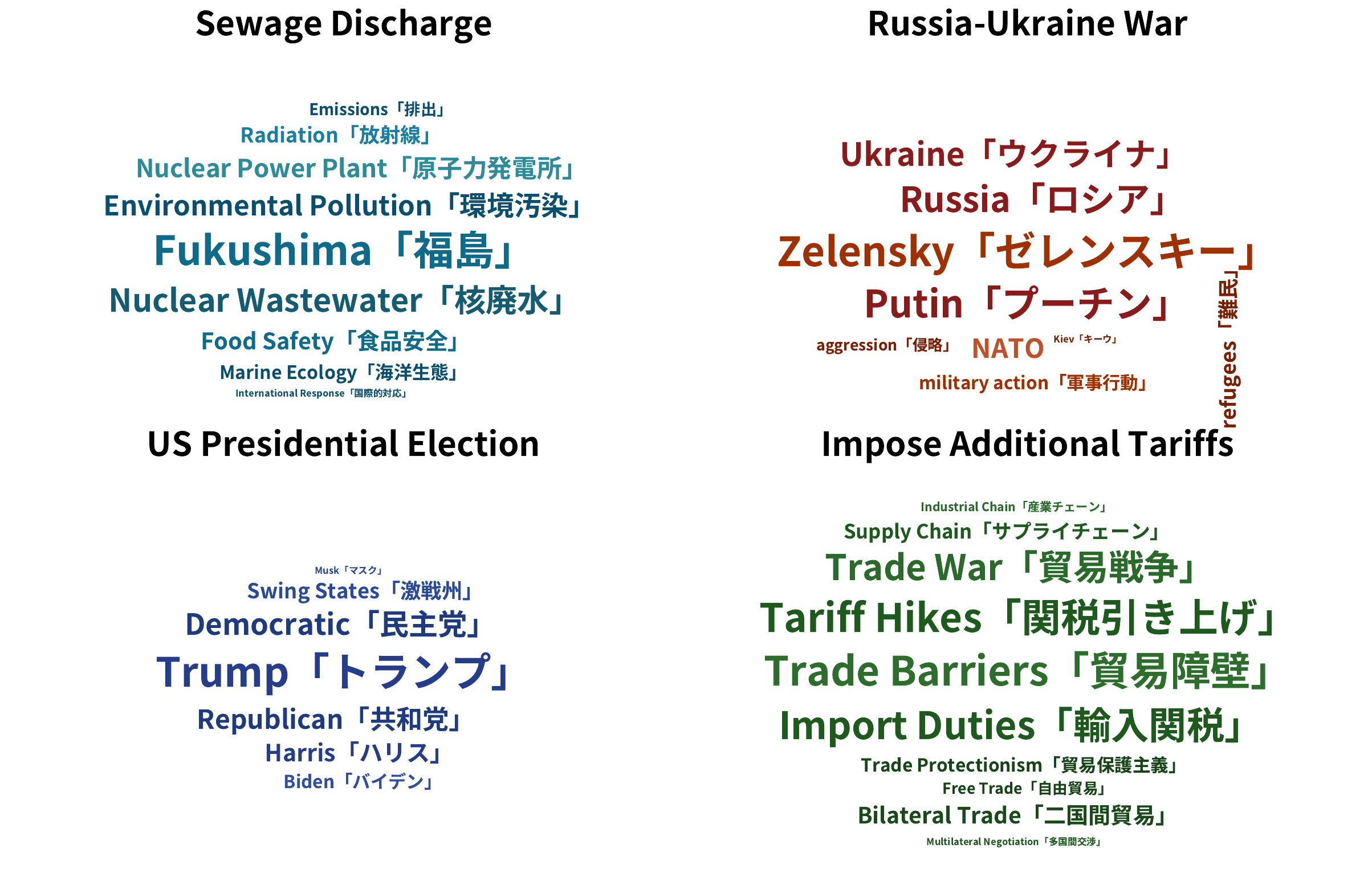}
		\caption{KIRP Dataset topic keywords}
		\label{fig:wordcloud}
	\end{figure}
	
	Our data were collected from Twitter, targeting events that triggered intense public discussion. We crawled tweets across four domains—social, political, diplomatic, and economic—covering the following topics: wastewater discharge, the U.S. presidential election, the Russia–Ukraine war, and tariff increases. The specific topics and their corresponding collection keywords are detailed in Fig. 1.
	
	This multi-domain design ensures that the collected dataset encompasses a wide range of stance targets, linguistic styles, and user perspectives, thereby enhancing the generalizability of our experimental evaluation.
	\subsection{Dataset Annotation}
	For the annotation process, we manually labeled data into three primary categories: Favor, Against, and Neutral. The Irrelevant labels were subsequently generated through target-data recombination. Our methodology follows the four-category scheme proposed by Conforti et al. ~\cite{conforti2020will}: Favor, Against, Neutral, and Irrelevant. We employed a three-party annotation process and utilized Fleiss’ Kappa coefficient to assess inter-annotator agreement. The specific annotation and filtering guidelines are defined as follows:
	\begin{itemize}
		\item For each topic$I$, annotations must ensure that $K_I\ge 0.9$ and the final dataset $K_D\ge 0.9$
		\item Data whose content is too simplistic to determine stance polarity or topic relevance are discarded.
		\item If a data instance criticizes or endorses both sides of the target stance, it is labeled as $Neutral$.
		\item If a data instance clearly expresses stance polarity but is unrelated to the target (e.g., Text: \textit{“Imposing additional tariffs is a necessary measure to address trade imbalance, protect domestic industries and employment, and serve national interests,”} Target: \textit{“Support Ukraine’s victory”}), it is assigned the $Irrelevant (I)$ label. Its associated domain and topic information remain unchanged, and such instances are excluded from Fleiss’ Kappa calculations.
	\end{itemize}
	
	Using this methodology, we constructed the KIRP dataset, which contains the following information in TABLE III.
	
	\subsection{Validity of Irrelevant Label Generation}
	To verify that the target-data recombination method produces irrelevant samples consistent with real-world distributions, we conducted three analyses. 
	First, three annotators independently judged 200 recombined instances against 200 naturally collected irrelevant tweets; the average agreement rate was 91.3\% (Fleiss’ $\kappa=0.87$). 
	Second, we compared TF-IDF and POS-tag distributions between the two sets, obtaining a cosine similarity of 0.86 and JS-divergence of 0.11, indicating high distributional overlap. 
	Third, a BERT-based binary classifier could not distinguish recombined from natural irrelevant samples above chance (54.2\% accuracy). 
	These results confirm that our generation method does not introduce artificial biases and serves as a valid proxy for real irrelevant stance data.
	
	
	
	\definecolor{headerbg}{RGB}{91, 132, 177}   
	\definecolor{rowB}{RGB}{245, 247, 250}
	\definecolor{rowA}{RGB}{255, 255, 255}
	\begin{table}[H]
		\centering
		\footnotesize
		\caption{Statistics of the KIRP Dataset, including the number
			(\textit{n}) and proportion (\%) of tweets per stance
			label across all four topics.}
		\label{tab:kirp_dataset}
		\renewcommand{\arraystretch}{1.25}
		\setlength{\tabcolsep}{3pt}
		\arrayrulecolor{black}   
		\begin{tabular*}{\columnwidth}{@{\extracolsep{\fill}}
				|l|r|r|r|r|r|r|r|r|}
			\hline
			\rowcolor{headerbg}
			\multicolumn{1}{|l|}{\color{white}\textbf{Topic}}
			& \multicolumn{2}{c|}{\color{white}\textbf{Favor}}
			& \multicolumn{2}{c|}{\color{white}\textbf{Against}}
			& \multicolumn{2}{c|}{\color{white}\textbf{Neutral}}
			& \multicolumn{2}{c|}{\color{white}\textbf{Total}} \\
			\hline
			\rowcolor{headerbg}
			\multicolumn{1}{|l|}{}
			& \multicolumn{1}{c|}{\color{white}\textit{n}}
			& \multicolumn{1}{c|}{\color{white}\textit{\%}}
			& \multicolumn{1}{c|}{\color{white}\textit{n}}
			& \multicolumn{1}{c|}{\color{white}\textit{\%}}
			& \multicolumn{1}{c|}{\color{white}\textit{n}}
			& \multicolumn{1}{c|}{\color{white}\textit{\%}}
			& \multicolumn{1}{c|}{\color{white}\textit{n}}
			& \multicolumn{1}{c|}{\color{white}\textit{\%}} \\
			\hline
			%
			\rowcolor{rowA}
			\makecell[l]{Sewage\\Discharge}
			& 4,256 & 40.6 & 3,121 & 29.8 & 3,103 & 29.6
			& \textbf{10,480} & \textbf{100} \\ \hline
			\rowcolor{rowA}
			\makecell[l]{Russia-\\Ukraine War}
			& 4,188 & 36.6 & 4,147 & 36.2 & 3,111 & 27.2
			& \textbf{11,446} & \textbf{100} \\ \hline
			\rowcolor{rowA}
			\makecell[l]{US Presidential\\Election}
			& 4,237 & 33.9 & 4,134 & 33.1 & 4,121 & 33.0
			& \textbf{12,492} & \textbf{100} \\ \hline
			\rowcolor{rowA}
			\makecell[l]{Impose Add.\\Tariffs}
			& 3,247 & 34.2 & 3,102 & 32.7 & 3,142 & 33.1
			& \textbf{9,491}  & \textbf{100} \\ \hline
			\rowcolor{rowA}
			\textbf{Overall}
			& \textbf{15,928} & \textbf{36.2}
			& \textbf{14,504} & \textbf{33.0}
			& \textbf{13,477} & \textbf{30.7}
			& \textbf{43,909} & \textbf{100} \\ \hline
		\end{tabular*}
		\arrayrulecolor{black}
	\end{table}
	
	\section{METHODOLOGY}
	Based on the discussed technical foundations, we designed a zero-shot, tweet-level stance detection model, as illustrated in Fig. 3. The architecture consists of three core components: (1) external knowledge injection and entity reorganization augmentation; (2) LLM-based reflective chain-of-thought (CoT) reasoning; and (3) stance contrastive learning with prototypical network classification.
	
	\begin{figure*}[h]
		\centering
		\includegraphics[width=0.9\textwidth]{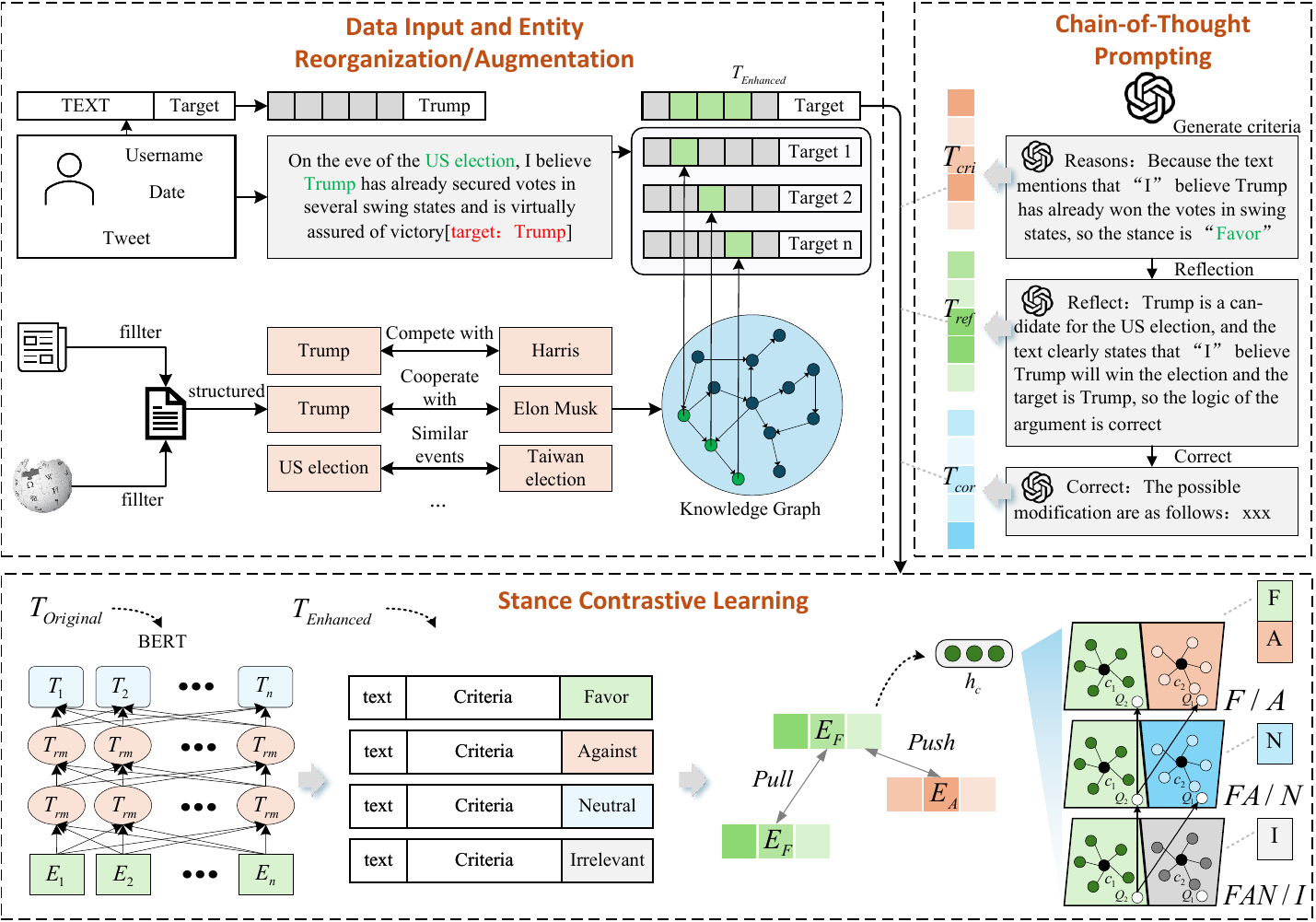} 
		\caption{\centering Architecture Diagram of the KIRP Zero-Shot Tweet-Level Stance Detection Model}
		\label{fig:p1}
	\end{figure*}
	
	\subsection{Problem Definition}
	We define the zero-shot stance detection task as follows. 
	Given a test dataset $D_s = \{(x_s^i, t_s^i, y_s^i)\}_{i=1}^{N_s}$ containing tweet texts, targets, and stance labels to be detected, where $x_s^i$ denotes the tweet text, $t_s^i$ represents the target corresponding to the tweet text, $y_s^i$ indicates the corresponding stance value, and $N_s$ denotes the dataset size. 
	We train a zero-shot stance detection model $f$ to compute, for each test pair $\langle x_s^i, t_s^i \rangle$ in $D_s$:
	
	\begin{equation}
		f(x_s^i, t_s^i) = y \in \{F, A, N, I\},
	\end{equation}
	
	where the value range of $y_s^i$ is the set $\{F, A, N, I\}$, representing \textbf{Favor} (support), \textbf{Against} (oppose), \textbf{Neutral}, and \textbf{Irrelevant} stance orientations of the given tweet text toward the target, respectively.
	
	\subsection{Entity Reorganization and Augmentation}
	We utilize Wikipedia and relevant news reports as external knowledge sources. By identifying key entities and their interrelationships, we extract seed terms and construct triples to form a structured knowledge graph, denoted as $G$. For each input instance, a localized subgraph $G'$ is extracted from $G$ by retrieving the $k$-hop neighborhood of entities mentioned in the input text. Within $G'$, entities that share the same relation type with a common neighbor are identified as equivalent entities and provided to the entity reorganization module for substitution. A graph autoencoder is then employed on $G'$ to learn entity vector representations and reconstruct potential missing links within the subgraph.
	
	The graph autoencoder comprises an encoder and a decoder. Specifically, a Graph Attention Network (GAT) first encodes the entities within the knowledge subgraph, capturing the influence of diverse edge relationships. These encoded representations are subsequently fed into a DistMult scoring function to transform entities and relations into low-dimensional vectors while preserving the graph's structural integrity.
	
	Specifically, in the encoder layer, for the $l$-th layer, the representation of a node can be expressed as follows:
	
	\begin{equation}
		h_i^{(l+1)} = \text{ELU}\left(\sum_{j \in N_i} \theta_j^{(l)} W_j^{(l)} h_j^{(l)}\right)
		\label{eq:encoder},
	\end{equation}
	
	where $W^{(l)}$ is the weight matrix of the $l$-th layer, mapping the node features from the previous dimension to the next dimension, $N_i$ is the set of neighbor nodes of node $i$, and $\theta_j^{(l)}$ is the attention coefficient dynamically computed based on nodes $j$ and $i$.
	
	The vectors encoded by GATs are then passed to the DistMult scoring function for decoding and reconstructing the links in the graph (during the encoding process, the relational encoding is discarded, which to some extent results in the loss of edge information). DistMult is a link prediction model based on matrix factorization. Its core idea is to compute the score of a triple through low-dimensional embeddings, with the formula as follows:
	
	\begin{equation}
		s(v_p, r, v_q) = \text{sigmoid}(h_{v}^T W_r h_{q})
		\label{eq:distmult},
	\end{equation}
	
	where $h_{v}$ and $h_{q}$ are the vector representations of the entities, $W_r$ is the vector representation of relation $r$ in the form of a diagonal matrix, and the function $s(v_p, r, v_q)$ denotes the score of the triple. This score is used to measure the likelihood of a connection between entities $v_p$ and $v_q$ via relation $r$.
	
	After reconstructing the edge information, the graph autoencoder is trained using the EdgeLoss module to ensure that the encoder and decoder can effectively reconstruct the links in the subgraph $G'$. EdgeLoss computes the difference between reconstructed triples and actual triples by comparing positive samples (i.e., connections that exist in $G'$) and negative samples (connections predicted by DistMult but absent from $G'$) to evaluate the loss of edge information. This method is implemented by training a binary cross-entropy loss function. For the subgraph $G'$, its EdgeLoss is expressed as:
	
	\begin{equation}
		\begin{split}
			L_{G'} = -\frac{1}{2|E'|} \sum_{(v_p, v_q, v_r) \in Z} \big( & y \log s(v_p, r, v_q) \\
			& + (1-y) \log(1-s(v_p, r, v_q)) \big)
		\end{split}
		\label{eq:edgeloss},
	\end{equation}
	
	where $E'$ is the set of edges in the subgraph $G'$, $Z$ represents the set of positive and negative samples (positive samples correspond to edges that exist in $G'$, negative samples correspond to edges absent from $G'$ but predicted by DistMult), and $y$ is an indicator variable that takes the value 1 for positive samples and 0 for negative samples.
	
	Finally, the GAT-encoded entity representations from subgraph $G'$ are aggregated to obtain a vector containing graph-aware features: $h_x = \{h_1, h_2, \dots, h_n\}$. These features encode both the structural information from $G'$ and the semantic knowledge inherited from the original knowledge graph $G$.
	
	We pair the tweet to be detected with its corresponding target to form a $\langle\text{text}, \text{target}\rangle$ input pair as the original input. In the entity reorganization and supplementation part, we extract key entities from the text using a large language model. We define key entities as those that can influence the final stance judgment, where key entities can only be nouns (such as specific event-related entities like ``Trump'' or ``US Election'') rather than verbs (such as expressions indicating actions like ``support'' or ``oppose''). For each extracted key entity, we query the subgraph $G'$ (extracted from the original knowledge graph $G$) to identify its equivalent entities. Two entities in $G'$ are considered equivalent if they share the same relation type with a common neighbor entity (e.g., $\langle\text{golf course}, \text{instance\_of}, \text{sports facility}\rangle$ and $\langle\text{football field}, \text{instance\_of}, \text{sports facility}\rangle$). By substituting the original key entity in the text with each identified equivalent entity, we generate multiple augmented $\langle\text{text}, \text{target}\rangle$ pairs from each original input, effectively expanding the training data with semantically diverse but structurally aligned variants.
	
	This substitution process generates multiple augmented detection pairs from the original text and target. These newly synthesized pairs are subsequently utilized to train the reflective chain-of-thought (CoT) module in the following stage.
	
	\subsection{Chain-of-Thought Prompting}
	We propose a new prompting chain—the Reflective Chain-of-Thought. Following the ToC method, we assign four hypothetical stance values \( S_p \) (support, oppose, neutral, irrelevant) to each generated data instance, and let the large language model analyze the reasonable explanations for each <text, target> pair under the hypothetical stance value \( S_p \) (even if this explanation may be incorrect). In the first stage, we force the large language model to generate such “counterfactual explanations”, obtaining a series of counterfactual arguments. Then in the second stage, we let the LLM further reflect on the generated counterfactual explanations, considering the logical flaws within them, thereby judging whether the hypothetical stance value \( S_p \) is correct. Different from ToC, to generate richer semantic information, we innovatively propose a third stage, where we let the large language model replace key entities in the original text to make the hypothetical stance value \( S_p \) valid, thereby enhancing the attention to key entities.
	
	Specifically, we design the following prompts and process in Fig. 3.
	\begin{figure}[!t]
		\centering
		\includegraphics[width=0.8\linewidth]{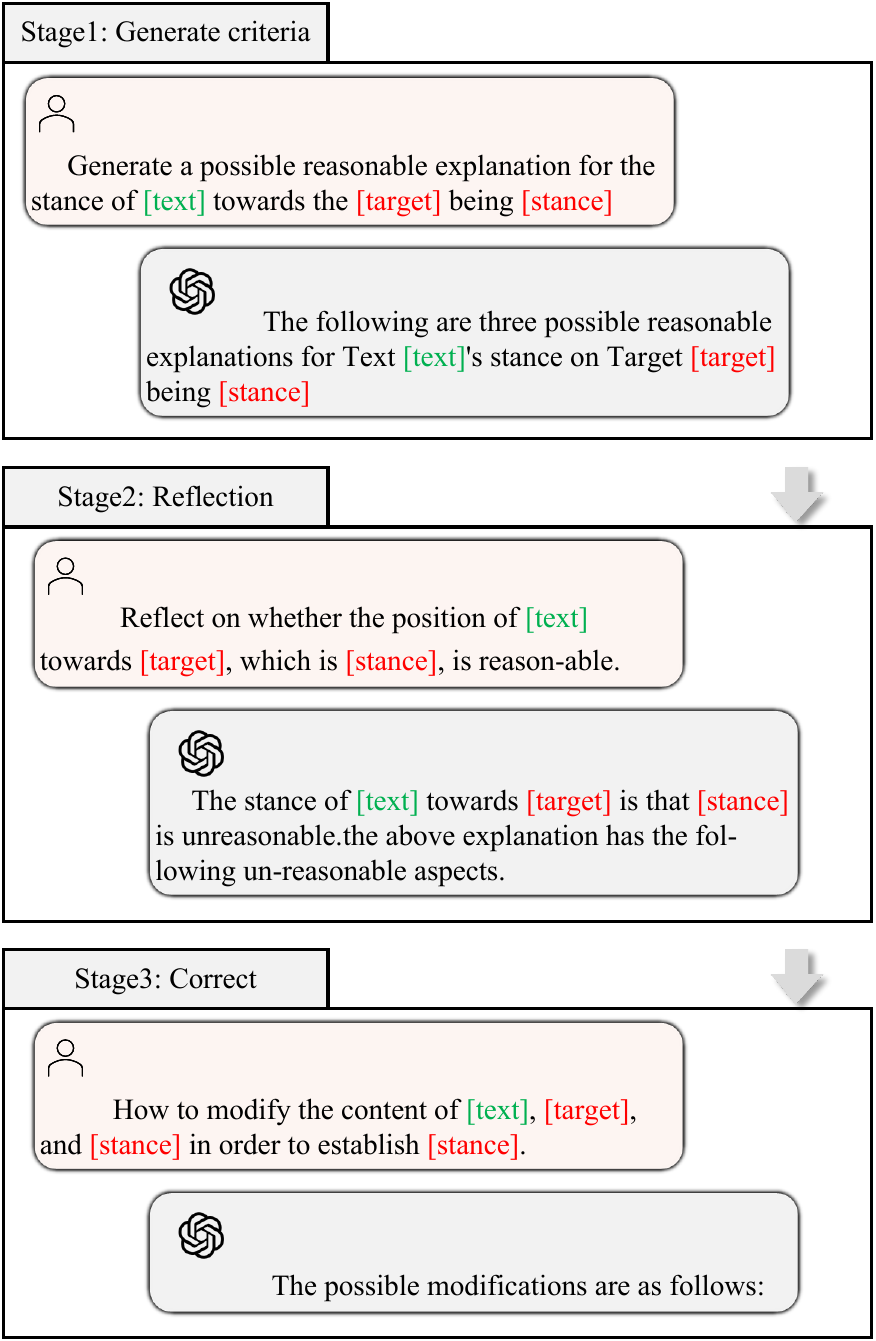}
		\caption{Prompt Example}
		\label{fig:prompt}
	\end{figure}
	For each input <text, target> detection pair, the three stages described above generate three text outputs \( T_{cri} \), \( T_{ref} \), and \( T_{cor} \). Among them, \( T_{cri} \) serves as supplementary criteria provided by the large language model and is directly added to the original text as supporting information, providing evidence to support the stance judgment for the brief tweet text. \( T_{ref} \) and \( T_{cor} \) are combined to form the stance label \( S_L \) generated by the large language model for the detection pair, which will be used in subsequent small model training.
	
	Finally, we combine the various possibilities obtained from the three stages to form multiple pieces of concatenated data in the format \textbf{[original data|criteria|stance]}, which are then input into the next layer of the model. This data contains rich contextual semantics and logical explanations, facilitating better extraction of textual features by the small model.
	
	\subsection{Stance Contrastive Learning}
	We concatenate the multiple stance representations generated from one data instance with the argument \( T_{ei} \) and encode them, then input them into the stance contrastive learning module for training. BERT embeds the obtained token input \( E = \{E_1, E_2, \dots, E_n\} \), converting each token value \( E_i \) into a word embedding vector by looking up the pre-trained embedding matrix, while adding positional embeddings to each word embedding. These embedding vectors are accumulated one by one to produce the final input embedding, which is provided to the BERT encoding layer.
	
	In BERT's multi-layer Transformer encoding layers, after the following four processes, the preliminary output vector \( h_z = \{h_1, h_2, \dots, h_n\} \) is obtained:
	
	\begin{enumerate}
		
		\item \textbf{Multi-Head Attention Mechanism:} Each Encoder and Decoder of the Transformer contains this component, allowing the model to compute the relationships between each token within the sequence and all other tokens. Through this mechanism, the model can capture long-range dependencies within the sequence.
		
		\item \textbf{Normalization and Residual Connection:} Immediately following the attention layer, a residual connection and layer normalization are applied. These two components are also present in each Encoder and Decoder of the Transformer, helping to avoid the vanishing gradient problem when training deep networks and promoting information flow between layers.
		
		\item \textbf{Feed-Forward Network:} The feed-forward network performs a nonlinear transformation on each token. In the first linear layer, the feed-forward network maps the input token from the original 768 dimensions to a higher-dimensional space, typically 3,072 dimensions. This operation is represented as:
		
		\begin{equation}
			FFN_1(x) = W_1x + b_1
			\label{eq:ffn1},
		\end{equation}
		
		After obtaining the high-dimensional output, ReLU is used as the activation function to introduce nonlinear factors. This process is represented as:
		
		\begin{equation}
			Activation(x) = \max(0, x)
			\label{eq:relu},
		\end{equation}
		
		In the second linear layer, the high-dimensional vector representation is mapped back to the original dimension:
		
		\begin{equation}
			FFN_2(x) = W_2\,Activation(FFN_1(x) + b_2)
			\label{eq:ffn2},
		\end{equation}
		
		The nonlinear transformation of the feed-forward network enables tokens to obtain richer contextual representations and strengthens the representation of thematic information in the vectors.
		
		\item \textbf{Normalization and Residual Connection Again:} Same as 2), to avoid gradient vanishing, normalization and residual connections are introduced again to process the vectors.
	\end{enumerate}
	
	After the BERT encoding layer, the preliminary output vector \( h_z = \{h_1, h_2, \dots, h_n\} \) with contextual information and thematic representations is obtained, providing input for subsequent stance contrastive learning.
	
	Stance contrastive learning enhances the model's generalization ability for stance features by comparing samples of the same stance category with samples of different stances. Specifically, given the hidden vectors \(\{h_i\}_{i=1}^N\) in a mini-batch, where \(N_b\) is the batch size and \(h_i\) is the hidden representation of the \(i\)-th vector. For each sample \(h_i\), samples \(h_j\) with the same stance value as this sample are called positive samples, while other samples \(h_k\) are called negative samples.
	
	The objective of stance contrastive learning is to train a loss function as follows:
	
	\begin{equation}
		L_s = -\frac{1}{N_b} \sum_{h_i \in B} c^s(h_i)
		\label{eq:stance_loss},
	\end{equation}
	
	where \(c^s(h_i)\) is the contrastive loss for sample \(h_i\), defined as:
	
	\begin{equation}
		c^s(h_i) = \log \frac{\sum_{j \in B \setminus \{i\}} \mathbf{1}_{y^i = y^j} \exp(f(h_i, h_j) / \tau_s)}{\sum_{j \in B \setminus \{i\}} \exp(f(h_i, h_j) / \tau_s)}
		\label{eq:contrastive_loss},
	\end{equation}
	
	where \(\mathbf{1}_{y^i = y^j}\) is an indicator function that takes the value 1 when \(y^i = y^j\), indicating that only samples with the same stance tendency as \(h_i\) are counted as positive samples.
	
	\(f(h_i, h_j)\) is the cosine similarity function, used to measure the similarity between two hidden vectors, defined as follows:
	
	\begin{equation}
		f(h_i, h_j) = \frac{h_i^T h_j}{\|h_i\| \|h_j\|}
		\label{eq:cosine_similarity},
	\end{equation}
	
	\(\tau_s\) is the temperature parameter, used to adjust the sensitivity of the contrastive loss. Ultimately, the goal of this loss function is to make samples of the same stance category closer in the feature space, while samples of different stance categories are farther apart. In this way, the model can better generalize to unseen targets, as similar stance features can be shared across different targets.
	
	\subsection{Iterative Prototype Network Classification}
	To achieve four-class stance classification (support, oppose, neutral, irrelevant), this paper trains three 2-way K-shot prototypical networks, implementing four-class stance classification through iterative binary classification. Specifically, the first-layer prototypical network first distinguishes whether a piece of data is topic-related; if it is topic-irrelevant, it directly outputs “irrelevant”. Otherwise, the query point is placed into the second-layer prototypical network to distinguish whether the query point has stance polarity; if it has no stance polarity, it outputs “neutral”. Otherwise, the query point is placed into the third-layer prototypical network to distinguish whether the query point has positive or negative polarity, correspondingly outputting “support” and “oppose” stances.
	
	The naming of each layer's prototypical network comes from the source datasets of the support set and query set used to train that prototypical network. Taking the first-layer prototypical network FAN/I as an example, the dataset set FAN (containing support, oppose, and neutral) and the dataset set I (containing irrelevant) are divided into multiple support sets \(S\) and query sets \(Q\), and the prototypical network is trained in episodes. The support set \(S\) and query set \(Q\) satisfy the following division:
	
	\begin{equation}
		S = \{(T_i^s, t_i^s, y_i^s)\}, \quad i \in [1, |S|]
		\label{eq:support_set},
	\end{equation}
	
	\begin{equation}
		Q = \{(T_j^q, t_j^q, y_j^q)\}, \quad j \in [1, |Q|]
		\label{eq:query_set},
	\end{equation}
	
	where the triplet \(\{T, t, y\}\) represents topic, text, and stance respectively, \(|S|\) and \(|Q|\) represent the sizes of the support set and query set respectively, and the above formulas additionally satisfy the condition that \(S\) and \(Q\) are disjoint.
	
	In one episode, let the input data \(x\) have obtained representation \(f(x)\) through the relevant model:
	
	\begin{equation}
		f(x) = \text{Model}(x)
		\label{eq:feature_representation},
	\end{equation}
	
	Calculate the prototype point \(P_c\) for each class, represented as the mean of all sample features in the support set within one episode, i.e., the center point of the feature space:
	
	\begin{equation}
		P_c = \frac{1}{N_c} \sum_{x \in S_c} f(x)
		\label{eq:prototype},
	\end{equation}
	
	where \(S_c\) is the set of samples of class \(c\) in the support set, and \(N_c\) is the number of samples in \(S_c\).
	
	For each sample in the query set, calculate the distance between its feature representation and each prototype, and then classify the sample into the feature space of the nearest prototype point:
	
	\begin{equation}
		\hat{y} = \arg \min_c d(f(x), P_c)
		\label{eq:classification},
	\end{equation}
	
	where \(\hat{y}\) is the predicted result, and \(d\) is the distance metric function.
	
	After completing the classification of all query points, a loss function is trained to evaluate the accuracy of the current episode and optimize the model parameters through backpropagation:
	
	\begin{equation}
		L = -\sum_{c=1}^C y_c \log\left(\text{softmax}(-d(f(x), P_c))\right)
		\label{eq:loss},
	\end{equation}
	
	where \(y_c\) is the one-hot encoding of the true class, and \(C\) is the total number of classes.
	
	This completes the training of one episode. Multiple similar episodes are repeated until the network loss falls below a set threshold.

	\section{RESULTS AND EVALUATION}
	The experiments in this paper are conducted on the following three datasets:
	
	\begin{enumerate}
		\item \textbf{SemEval 2016 T6}: This dataset is derived from Twitter texts and covers four topics: Feminist Movement (FM), Legalization of Abortion (LA), Hillary Clinton (HC), and Donald Trump (DT). Based on the text content, the dataset annotators set the stance labels as three categories: support, oppose, and neutral.
		
		\item \textbf{WT-WT}: This is one of the largest stance detection datasets to date. Its main content concerns the interaction issues between eight major companies discussed on Twitter. Different companies have distinct attributes, and the keywords they involve also vary. WT-WT classifies the statements describing inter-company operations and the attitudes held toward these operations into four categories: Support, Refute, Comment, and Unrelated.
		
		\item \textbf{KIRP-D}: The dataset proposed in this paper covers four domains (Society, Politics, Diplomacy, Economy).
	\end{enumerate}
	
	\subsection{Validity of Method}
	To evaluate the performance of the external knowledge-enhanced few-shot tweet-level stance detection model, this section first applies the model to conventional stance detection tasks, comparing 23 baseline models on the above three datasets, including:
	\begin{table*}[!h]
		\centering
		\small
		\caption{Results on KIRP-D, SemEval-2016, and WT-WT}
		\label{tab:semeval2}
		\begin{tblr}{
				colspec  = {X[2.0,l,m]
					|X[0.4,c,m]X[0.4,c,m]X[0.4,c,m]X[0.4,c,m]X[0.5,c,m]
					|X[0.4,c,m]X[0.4,c,m]X[0.4,c,m]X[0.4,c,m]X[0.4,c,m]X[0.4,c,m]X[0.5,c,m]
					|X[0.4,c,m]X[0.4,c,m]X[0.4,c,m]X[0.4,c,m]X[0.5,c,m]},
				width    = \textwidth,
				rowsep   = 1pt,
				row{4,9,15,19,25,32} = {bg=gray!15},
				hline{1}  = {-}{1.6pt},
				hline{2}  = {-}{0.4pt},
				hline{3}  = {-}{0.4pt},
				hline{4}  = {-}{0.4pt},
				hline{5}  = {-}{0.4pt},
				hline{9}  = {-}{0.4pt},
				hline{10} = {-}{0.4pt},
				hline{15} = {-}{0.4pt},
				hline{16} = {-}{0.4pt},
				hline{19} = {-}{0.4pt},
				hline{20} = {-}{0.4pt},
				hline{25} = {-}{0.4pt},
				hline{26} = {-}{0.4pt},
				hline{32} = {-}{0.4pt},
				hline{33} = {-}{0.4pt},
				hline{34} = {-}{1.6pt},
				cell{1}{1}  = {r=3}{c,m},
				cell{1}{2}  = {c=17}{c,m},
				cell{2}{2}  = {c=5}{c,m},
				cell{2}{7}  = {c=7}{c,m},
				cell{2}{14} = {c=5}{c,m},
				row{1,2,3}  = {font=\bfseries\small},
			}
			Model & F1-score & & & & & & & & & & & & & & & & \\
			& KIRP-D & & & & & SEM-16 & & & & & & & WT-WT & & & & \\
			& SD & RUW & USP & IAT & All & DT & HC & FM & LA & A & CC & All & CA & CE & AC & AH & All \\
			\textbf{BiLSTM-based} & & & & & & & & & & & & & & & & & \\
			BiCond~\cite{augenstein2016stance}      & 44.91 & 45.87 & 35.78 & 38.87  & 42.90 & 30.47 & 33.54 & 38.23 & 34.42 & 31.11 & 16.83 & 29.74 & 55.85 & 51.79 & 63.58 & 61.75 & 56.61 \\
			CrossNet~\cite{xu2018cross}             & 46.45 & 45.08 & 39.29 & 37.26 & 42.51 & 35.58 & 40.25 & 41.70 & 37.77 & 39.65 & 22.75 & 36.89 & 57.93 & 53.77 & 64.80 & 59.18 & 58.45 \\
			TAN~\cite{du2017stance}                 & 50.29 & 47.06 & 38.04 & 44.72  & 44.62 & 40.01 & 43.67 & 39.91 & 41.10 & 42.28 & 28.74 & 37.65 & 51.72 & 48.31 & 38.89 & 41.57 & 44.49 \\
			SKET~\cite{zhang2020enhancing}          & 50.36 & 46.58 & 37.81 & 32.42 & 42.29 & --- & --- & --- & --- & --- & --- & --- & --- & --- & --- & --- & --- \\
			\textbf{BERT-based} & & & & & & & & & & & & & & & & & \\
			BERTweet                                & \uline{69.04} & \uline{66.98} & 77.67 & 74.52 & 71.15 & 52.75 & 46.37 & 48.09 & 50.01 & 45.57 & 47.87 & 47.26 & 62.73 & 58.74 & 56.38 & 60.19 & 59.21 \\
			TGA-Net*~\cite{allaway2020zero}          & 52.68 & 55.52 & 79.82 & 67.77 & 63.18 & 40.68 & 50.21 & 46.64 & 45.17 & 52.67 & 36.66 & 42.78 & 64.70 & 63.48 & 67.25 & 67.57 & 66.61 \\
			TGA-Net-ft*~\cite{allaway2020zero}       & 54.50 & 56.56 & 86.27 & 67.66 & 66.72 & 41.72 & 50.24 & 45.78 & 47.83 & 52.74 & 37.25 & 44.48 & 65.13 & 64.27 & 66.83 & 68.98 & 67.04 \\
			RoBERTa-Enriched                        & 58.55 & 59.52 & 76.19 & 72.11 & 66.23 & 47.24 & 50.87 & 48.93 & 56.25 & 55.24 & 49.68 & 51.57 & 61.56 & 64.38 & 62.64 & 61.78 & 62.00 \\
			ANEK*~\cite{chunling2023adversarial}     & 55.09 & 57.15 & 62.02 & 60.60 & 58.18 & 50.26 & 54.70 & 55.21 & 48.77 & 54.14 & 39.20 & 47.67 & 70.89 & 70.01 & 72.77 & 72.33 & 71.56 \\
			\textbf{EK-based} & & & & & & & & & & & & & & & & & \\
			MT-DNN~\cite{liu2019multi}              & 56.11 & 54.56 & 59.91 & 57.72   & 58.58 & 45.40 & 49.58 & 53.45 & 52.38 & 56.35 & 55.03 & 53.17 & 57.75 & 58.65 & 57.24 & 59.27 & 58.01 \\
			CKE-Net*~\cite{liu2021enhancing}         & 55.88 & 58.04 & 63.78 & 50.75 & 60.47 & 46.79 & 47.24 & 50.00 & 49.28 & 51.15 & 51.04 & 49.27 & 56.78 & 58.91 & 63.76 & 57.24 & 60.14 \\
			VTN~\cite{wei2019modeling}              & 57.67 & 62.90 & 66.99 & 48.54 & 63.72 & 49.25 & 50.10 & 47.94 & 51.88 & 49.35 & 47.67 & 49.24 & 59.02 & 64.02 & 58.87 & 62.44 & 62.45 \\
			\textbf{CL-based} & & & & & & & & & & & & & & & & & \\
			JointCL*~\cite{liang2022jointcl}         & 62.35 & 62.32 & 85.64 & 76.91 & 73.80 & 50.50 & 54.67 & 53.28 & 48.98 & 54.45 & 40.70 & 50.78 & 72.37 & 70.20 & 75.68 & 73.77 & 73.89 \\
			STCC~\cite{liu2022target}               & 54.04 & 53.67 & 75.47 & 60.06   & 64.45 & 46.24 & 48.90 & 51.25 & 47.93 & 52.25 & 49.01 & 49.57 & 62.78 & 65.12 & 64.84 & 62.09 & 63.77 \\
			DTCL*~\cite{liu2022connecting}           & 59.69 & 60.55 & 85.11 & 78.94 & 70.12 & 50.15 & 49.80 & 51.48 & 50.03 & 49.07 & 50.52 & 50.14 & 60.09 & 61.80 & 66.54 & 63.91 & 62.57 \\
			PT-HCL*~\cite{liang2022zero}             & 59.55 & 60.81 & \uline{87.13} & 69.60 & 72.86 & 49.98 & 54.35 & 54.60 & 48.79 & 54.78 & 51.24 & 53.21 & 73.17 & 69.25 & 75.57 & 76.11 & 73.75 \\
			MTFF*~\cite{zhao2024zero}                & 57.12 & 58.87 & 80.10 & 73.31 & 69.32 & 48.32 & 51.26 & 49.45 & 53.08 & 46.84 & 52.14 & 49.28 & 58.19 & 59.99 & 64.32 & 61.27 & 60.24 \\
			\textbf{LLM-based} & & & & & & & & & & & & & & & & & \\
			GPT-3.5-CoT*                             & 62.09 & 61.47 & 79.38 & 72.47 & 68.60 & 74.32 & 75.08 & 69.89 & 75.24 & 77.40 & 73.55 & 72.07 & 77.98 & 76.54 & 77.47 & 79.07 & 78.62 \\
			GPT-3.5-ToC*                             & 63.38 & 63.86 & 86.30 & 72.77 & 72.02 & 76.58 & 80.26 & 81.44 & 78.25 & 77.44 & 75.26 & 79.89 & 80.25 & 81.64 & 79.38 & 82.75 & 80.84 \\
			DS-ESD~\cite{ding2024distantly}         & 63.62 & 62.18 & 83.59 & 70.02 & 72.60 & 65.43 & 63.82 & 69.90 & 71.25 & 69.45 & 66.68 & 68.55 & 73.56 & 75.67 & 74.32 & 75.98 & 74.36 \\
			KASD-GPT*                                & 61.07 & 62.00 & 80.52 & 68.85 & 70.32 & 64.25 & 61.57 & 66.37 & 67.76 & 63.88 & 68.45 & 65.77 & 75.75 & 74.99 & 79.05 & 73.28 & 75.03 \\
			{DeepSeek-V4 \\ CoT*}                    & 63.09 & 62.77 & 82.75 & 74.02 & 71.90 & 80.32 & 81.22 & 79.58 & 84.89 & 82.20 & 78.58 & 81.74 & 80.07 & 81.25 & 79.57 & 83.44 & 81.29 \\
			{DeepSeek-V4 \\ ToC*}                    & 65.32 & 64.10 & 86.28 & \uline{74.92} & \uline{74.50} & \textbf{83.27} & \uline{83.28} & \uline{82.73} & \textbf{84.57} & \uline{84.39} & \uline{80.03} & \uline{83.91} & \uline{82.93} & \uline{82.25} & \textbf{83.98} & \uline{84.02} & \uline{83.31} \\
			\textbf{Our\_Model} & & & & & & & & & & & & & & & & & \\
			KIRP*                                    & \textbf{69.78} & \textbf{70.16} & \textbf{90.22} & \textbf{80.66} & \textbf{79.18} & \uline{82.28} & \textbf{85.06} & \textbf{83.17} & \uline{84.21} & \textbf{85.50} & \textbf{81.74} & \textbf{84.05} & \textbf{84.78} & \textbf{85.20} & \textbf{85.79} & \uline{83.98} & \textbf{84.99} \\
		\end{tblr}
	\end{table*}
	
	BiCond ~\cite{augenstein2016stance}, CrossNet ~\cite{xu2018cross}, TAN ~\cite{du2017stance}, and SKET ~\cite{zhang2020enhancing}, which leverage sequential encoding and attention mechanisms for stance detection. BERT-based models include BERTweet, TGA-Net, TGA-Net-ft ~\cite{allaway2020zero}, RoBERTa+Enriched, and ANEK ~\cite{chunling2023adversarial}, which exploit pre-trained language representations with varying levels of target awareness. External knowledge injection-based models include MT-DNN ~\cite{liu2019multi}, CKE-Net ~\cite{liu2021enhancing}, and VTN ~\cite{wei2019modeling}, which incorporate structured knowledge to improve cross-target generalization. Contrastive learning-based models include JointCL ~\cite{liang2022jointcl}, STCC ~\cite{liu2022target}, DTCL ~\cite{liu2022connecting}, PT-HCL ~\cite{liang2022zero}, and MTFF ~\cite{zhao2024zero}, which capture discriminative stance features through sample contrast. Finally, LLM-based models include GPT-3.5-CoT, GPT-3.5-ToC, DS-ESD ~\cite{ding2024distantly}, KASD-ChatGPT, DeepSeek-CoT, and DeepSeek-ToC, which leverage large-scale language model reasoning for zero-shot stance inference.

	The above 23 models were experimented on the three datasets. To ensure a fair comparison between data-driven models and zero-shot models, different evaluation settings were adopted. Data-driven models were trained using 100\% of the original training data, while zero-shot models (marked with *) were evaluated under the 0-shot setting. In Table~\ref{tab:semeval2}, the best results are shown in bold and the second-best results are underlined. From the results, we draw the following conclusions:
	
	First, \textbf{KIRP consistently achieves the highest overall F1 across all three datasets}. On KIRP-D, KIRP obtains 79.18 All-F1, outperforming the best LLM-based baseline DeepSeek-V4-ToC (74.50) by 4.68 points and the best contrastive learning model JointCL (73.80) by 5.38 points. On SEM-16, KIRP reaches 84.05 All-F1, surpassing DeepSeek-V4-ToC (83.91) by 0.14 points. On WT-WT, KIRP achieves 84.99 All-F1, leading DeepSeek-V4-ToC (83.31) by 1.68 points. Notably, KIRP is the only model that achieves top performance simultaneously on the Japanese four-class dataset (KIRP-D) and the two established English benchmarks, demonstrating the cross-lingual and cross-domain generalization of our framework.
	
	Second, \textbf{KIRP shows strong and consistent advantages across individual topic categories within each dataset}. On KIRP-D, KIRP leads all baselines across all four topics (SD, RUW, USP, IAT), with particularly notable margins on USP (90.22 vs.\ 87.13 by PT-HCL) and IAT (80.66 vs.\ 78.94 by DTCL). On SEM-16, KIRP achieves the best F1 on five of six topic categories (HC: 85.06, FM: 83.17, A: 85.50, CC: 81.74, All: 84.05), closely competing with DeepSeek-V4-ToC on the remaining categories. On WT-WT, KIRP leads on CA (84.78), CE (85.20), and AC (85.79), confirming that its advantages are not confined to specific topics but reflect a broadly superior representation capability.
	
	Third, \textbf{LLM-based prompting methods, especially ToC variants, significantly outperform traditional BiLSTM-, BERT-, and knowledge-enhanced models} across all three datasets. DeepSeek-V4-ToC (74.50 All-F1 on KIRP-D) surpasses the best EK-based model VTN (63.72) by 10.78 points and the best BERT-based model BERTweet (71.15) by 3.35 points. This highlights the effectiveness of counterfactual reasoning and self-reflection for stance detection in low-resource and non-English settings. However, KIRP further improves upon DeepSeek-V4-ToC across all three datasets, demonstrating that the combination of reflective CoT reasoning with prototype-network-based classification and stance-aware contrastive learning yields additional gains beyond pure LLM prompting.
	
	Fourth, \textbf{the performance gap between KIRP and LLM-based baselines is most pronounced on KIRP-D}, the Japanese four-class dataset (4.68 points over DeepSeek-V4-ToC), while the gaps narrow on the English benchmarks SEM-16 (0.14 points) and WT-WT (1.68 points). This pattern suggests that KIRP's external knowledge injection and entity reorganization provide the greatest benefit in the more challenging cross-lingual, data-scarce scenario. On the well-resourced English benchmarks, where LLM-based methods already perform strongly, KIRP still achieves state-of-the-art results, confirming the robustness of its framework across different resource conditions.
	
	In summary, the results validate that KIRP achieves state-of-the-art stance detection performance across all three benchmarks. Its advantages are consistent at both the dataset level and the individual topic level, and are most substantial in the challenging Japanese four-class setting, underscoring the effectiveness of integrating knowledge-based augmentation, reflective reasoning, and prototype classification for zero-shot tweet-level stance detection.
	
	\subsection{Ablation Study}
	\subsubsection{Component Ablation Experiment}
	To further analyze the contribution of each model component, this paper designs an ablation experiment. The models without stance contrastive learning (M$\backslash$CL), without the external knowledge graph (M$\backslash$KG), without the prototypical network (M$\backslash$PN), and without the large language model prompting chain (M$\backslash$LLM) are evaluated and analyzed in comparison with the complete model (M). The dataset used is KIRP-D, with the same evaluation metrics as in Experiment A. The prompt sample size for the large language model is 0-shots.
	
	\begin{table}[h]
		\centering
		\small
		\caption{KIRP Feature Ablation Experiment Results}
		\label{tab:ablation_component}
		\begin{tblr}{
				colspec  = {X[1.2,c,m] X[1,c,m] X[1,c,m] X[1,c,m] X[1,c,m]},
				width    = \linewidth,
				rowsep   = 2pt,
				hline{1} = {-}{1.2pt},
				hline{2} = {-}{0.4pt},
				hline{7} = {-}{1.2pt},
				row{1}   = {font=\bfseries\small},
			}
			Model & Accuracy & Precision & Recall & F1-score \\
			\textbf{M}    & \textbf{80.25} & \textbf{76.69} & \textbf{80.97} & \textbf{79.18} \\
			M$\backslash$CL  & 78.25 & 76.62 & 80.15 & 77.00 \\
			M$\backslash$KG  & 73.57 & 72.57 & 75.01 & 73.18 \\
			M$\backslash$PN  & 74.80 & 74.32 & 77.19 & 75.24 \\
			M$\backslash$LLM & \uline{65.74} & \uline{70.01} & \uline{71.14} & \uline{69.34} \\
		\end{tblr}
	\end{table}
	
	TABLE V shows the results of the ablation experiment. The experimental results indicate that stance contrastive learning, the external knowledge graph, the prototypical network, and the large language model prompting chain all play positive roles in the model's performance. Among these, M$\backslash$LLM performs the worst, demonstrating that the large language model's counterfactual prompting chain contributes the most to the model. The reason is that the data in the dataset mostly consist of brief Twitter social texts lacking contextual information and topic representation. The large language model provides rich contextual information for these short texts, offering criteria to support stance determination and logical reasoning, thereby enhancing the representation of textual features and topic information. Secondly, M$\backslash$PN performs relatively poorly, mainly reflected in the difficulty of accurately distinguishing between “irrelevant” and “neutral” stances using only fully connected layers and Softmax outputs. In contrast, the iterative prototypical network method outputs these two types of labels at different layers of the prototypical network, effectively distinguishing between these two highly similar labels. The component with the least impact is the stance contrastive learning module. The reason is that topic information has already been represented to a certain extent in the J-BERT model. The introduction of stance contrastive learning as a data enhancement module strengthens the vectors' discriminative power for different stance labels, contributing less to subsequent stance classification, yet it remains an effective method for improving model performance.
	
	\subsubsection{Prompting Chain Stage Ablation Experiment}To further analyze the contributions of the three stages in the reflective prompting chain, we designed experiments involving a prompting chain without criteria generation (LLM$\backslash$CRI), a prompting chain without reflection generation (LLM$\backslash$REF), and a prompting chain without correction generation (LLM$\backslash$COR). These were evaluated and analyzed in comparison with the complete reflective prompting chain (LLM) and the conventional reasoning prompting chain (CoT). Experiments were conducted using the KIRP dataset, and the results are as TABLE VI:
	
	\begin{table}[H]
		\centering
		\small
		\caption{Prompting Chain Stage Ablation Experiment Results}
		\label{tab:ablation_prompt}
		\begin{tblr}{
				colspec  = {X[1.2,c,m] X[1,c,m] X[1,c,m] X[1,c,m] X[1,c,m]},
				width    = \linewidth,
				rowsep   = 2pt,
				hline{1} = {-}{1.2pt},
				hline{2} = {-}{0.4pt},
				hline{7} = {-}{1.2pt},
				row{1}   = {font=\bfseries\small},
			}
			Model & Accuracy & Precision & Recall & F1-score \\
			\textbf{LLM}    & \textbf{80.25} & \textbf{76.69} & \textbf{80.97} & \textbf{79.18} \\
			LLM$\backslash$CRI  & \uline{68.14} & \uline{71.31} & \uline{73.56} & \uline{72.13} \\
			LLM$\backslash$REF  & 73.43 & 74.12 & 74.67 & 74.43 \\
			LLM$\backslash$COR  & 73.98 & 74.53 & 76.14 & 75.24 \\
			CoT & 75.88 & 74.78 & 75.90 & 75.00 \\
		\end{tblr}
	\end{table}
	
	TABLE VI shows the results of the ablation experiment on the reflective prompting chain stages. It can be observed that LLM$\backslash$CRI performs the worst, because the criteria generation stage provides rich contextual information for the brief tweet texts. Without the information supplement from this stage, the subsequent training of the small model is significantly affected. As supplements to the prompting chain, LLM$\backslash$REF and LLM$\backslash$COR also show a certain degree of performance degradation compared to the complete prompting chain. The reason is that these two stages supplement the model's reflection phase, promoting the large language model's self-thinking, reflection, and correction. The reflective prompting chain that integrates all three stages performs better overall than the conventional reasoning prompting chain CoT, effectively demonstrating the contribution of the reflective chain-of-thought in this model.
	
	In summary, the experimental results demonstrate that each component of the proposed framework—stance contrastive learning, external knowledge graph, prototypical network, and the reflective LLM prompting chain—contributes synergistically to the model's performance.Specifically, the reflective prompting chain proves to be the most critical module, as its absence (M$\backslash$LLM) leads to the most significant performance drop, emphasizing the necessity of deep semantic reasoning for understanding stance in short, informal texts. The external knowledge graph and prototypical network provide essential structural priors and stable classification boundaries, respectively, which are vital for zero-shot generalization. Furthermore, the contrastive learning module effectively refines the feature space, reducing intra-class variance.
	
	\subsection{Dataset Perturbation Attack Robustness Experiment}To verify the rationality of our dataset annotation and the robustness of our proposed model against perturbations, we conducted poisoning operations on 10\%, 30\%, and 50\% of the data sequentially across the three datasets: SemEval T6, WT-WT, and KIRP. The poisoning operations included generating adversarial samples through synonym substitution for a piece of data (e.g., replacing expressions in the text from “oppose” to “support”) while keeping the labels unchanged, and modifying the labels of a piece of data while keeping the content unchanged. We selected five better-performing methods from the five categories of baseline models—BiLSTM-based, BERT-based, EK-based, CL-based, and LLM-based—to compare with the KIRP model. The results are shown in TABLE VII.
	
	It can be observed that data-driven models are susceptible to perturbations from adversarial samples, while models based on large language model counterfactual prompting chains are less affected by adversarial samples due to their strong logical reasoning capabilities. Among these, the KIRP dataset exhibits the strongest adversarial perturbation against various models, with the highest rate of performance degradation, indicating that the KIRP data annotation is highly reasonable and the data accuracy is high. The KIRP model is relatively less affected by adversarial samples, which can be attributed to the enhancement provided by the ToC prompting chain. However, due to the presence of the contrastive learning module within the model, adversarial samples still interfere with the KIRP model's accuracy to some extent, making it slightly less robust than pure large language model reasoning chains.

	\begin{table}[H]
		\centering
		\small
		\caption{KIRP Dataset Perturbation Adversarial Attack Results}
		\label{tab:adversarial}
		\begin{tblr}{
				colspec  = {X[2.2,c,m] X[0.6,c,m]X[0.6,c,m]X[0.6,c,m]
					X[0.6,c,m]X[0.6,c,m]X[0.6,c,m]
					X[0.6,c,m]X[0.6,c,m]X[0.6,c,m]},
				width    = \linewidth,
				rowsep   = 2pt,
				hline{1} = {-}{1.2pt},
				hline{2} = {-}{0.4pt},
				hline{3} = {-}{0.4pt},
				hline{9} = {-}{1.2pt},
				cell{1}{1} = {r=2}{c,m},
				cell{1}{2} = {c=3}{c,m},
				cell{1}{5} = {c=3}{c,m},
				cell{1}{8} = {c=3}{c,m},
				row{1,2} = {font=\bfseries\small},
			}
			Model & WT-WT & & & SemEval T6 & & & KIRP & & \\
			& 10\%  & 30\% & 50\% & 10\% & 30\% & 50\% & 10\% & 30\% & 50\% \\
			SKET      & 40.7 & 31.5 & 27.7 & 43.8 & 35.4 & 28.1 & 39.4 & 30.6 & 21.8 \\
			BERTweet  & 68.7 & 65.3 & 45.9 & 70.8 & 62.5 & 46.6 & 69.3 & 63.7 & 47.5 \\
			VTN       & 60.8 & 55.3 & 41.7 & 63.7 & 56.7 & 45.2 & 59.7 & 47.2 & 40.6 \\
			JointCL   & 70.7 & 61.3 & 46.8 & 71.6 & 65.8 & 52.9 & 71.2 & 59.3 & 41.6 \\
			DS-R1-Toc & 73.7 & 72.8 & 73.2 & 75.5 & 74.4 & 74.1 & 77.8 & 76.4 & 76.4 \\
			KIRP      & 77.4 & 73.6 & 67.8 & 78.2 & 75.7 & 70.4 & 76.2 & 71.9 & 65.7 \\
		\end{tblr}
	\end{table}
	
	\begin{figure}[H]
		\centering
		\includegraphics[width=0.9\linewidth]{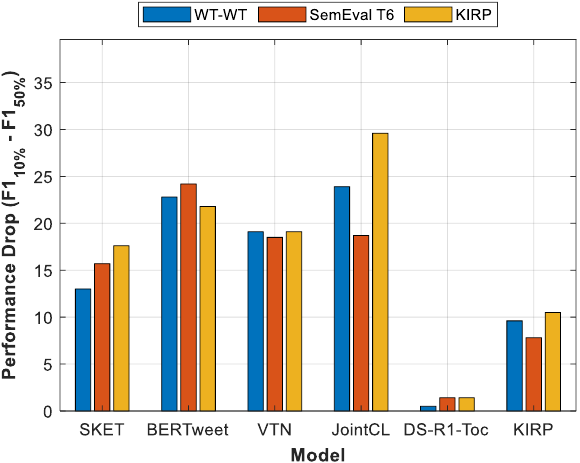}
		\caption{Comparison of Performance Degradation Values Across Models}
		\label{fig:degradation}
	\end{figure}
	
	This phenomenon stems from the architectural differences between the two models. DS-R1-ToC performs stance inference purely through counterfactual prompting chains, relying on the LLM's internal commonsense knowledge and logical reasoning rather than fitting to training data distributions. Adversarial perturbations therefore have limited impact, as its judgments are driven by semantic understanding instead of statistical patterns learned from labeled data. In contrast, KIRP's stance-aware contrastive learning module explicitly optimizes the representation space by pulling same-label samples closer and pushing different-label samples apart, which inherently depends on correct sample-label correspondences. When adversarial perturbations introduce mismatched text-label pairs, the contrastive objective may be misled into learning distorted feature boundaries, resulting in greater performance degradation relative to pure LLM reasoning chains.
	
	\subsection{Dataset Topic Drift Robustness Experiment}
	
	\begin{figure*}[h]
		\centering
		
		\begin{minipage}[c]{0.328\textwidth}
			\centering
			\includegraphics[width=\linewidth]{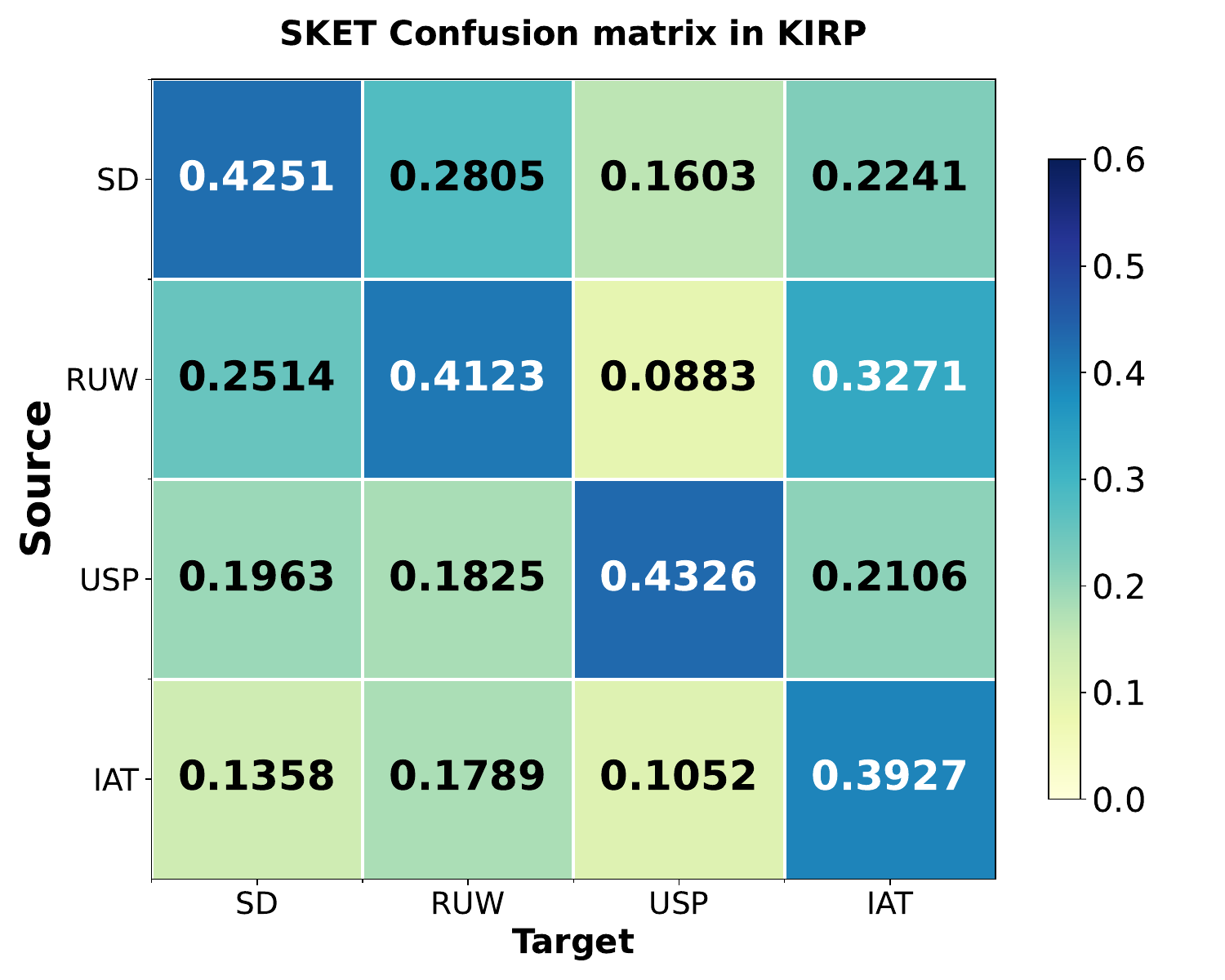}
			\caption*{(a)}
		\end{minipage}
		\hfill
		\begin{minipage}[c]{0.328\textwidth}
			\centering
			\includegraphics[width=\linewidth]{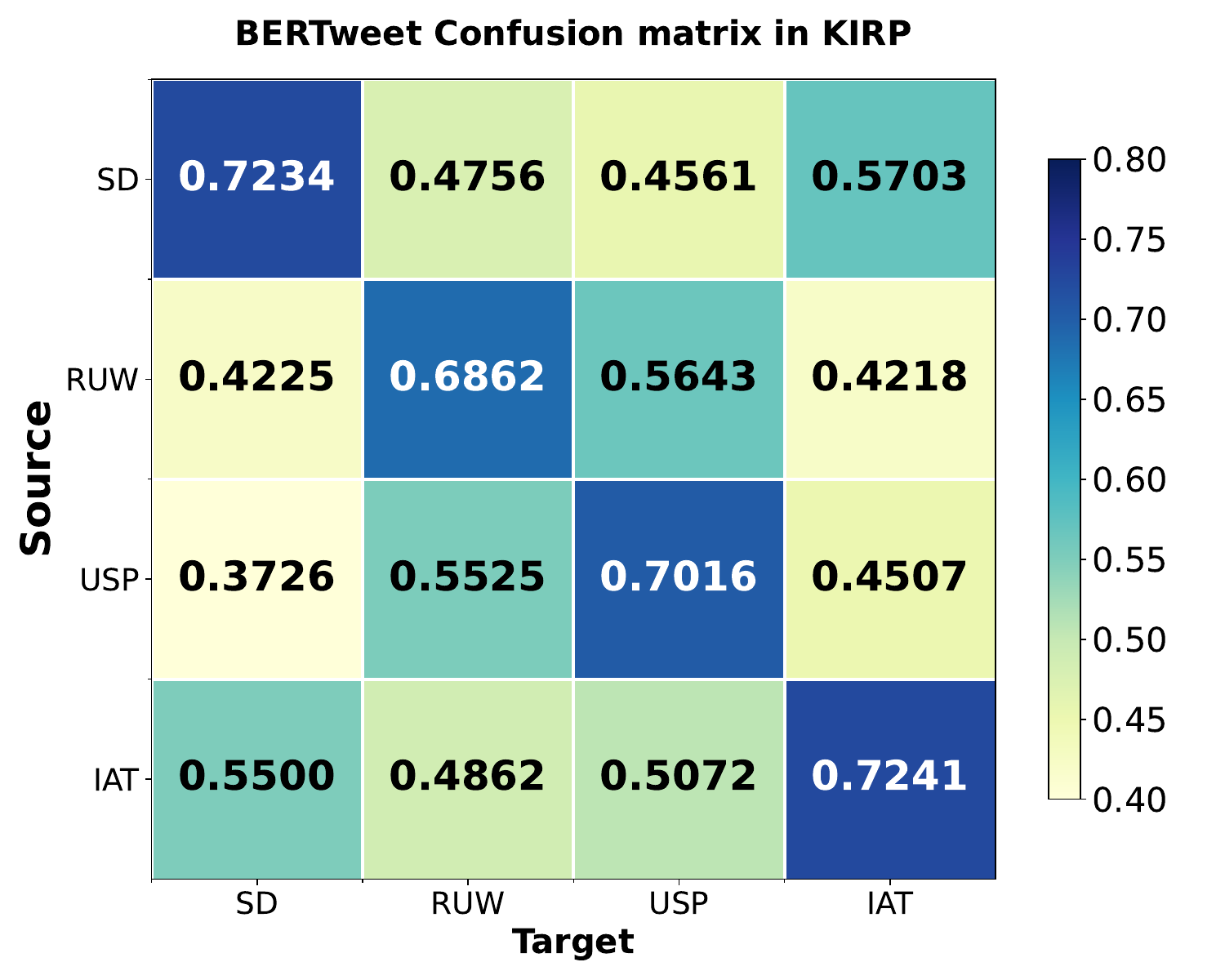}
			\caption*{(b)}
		\end{minipage}
		\hfill
		\begin{minipage}[c]{0.328\textwidth}
			\centering
			\includegraphics[width=\linewidth]{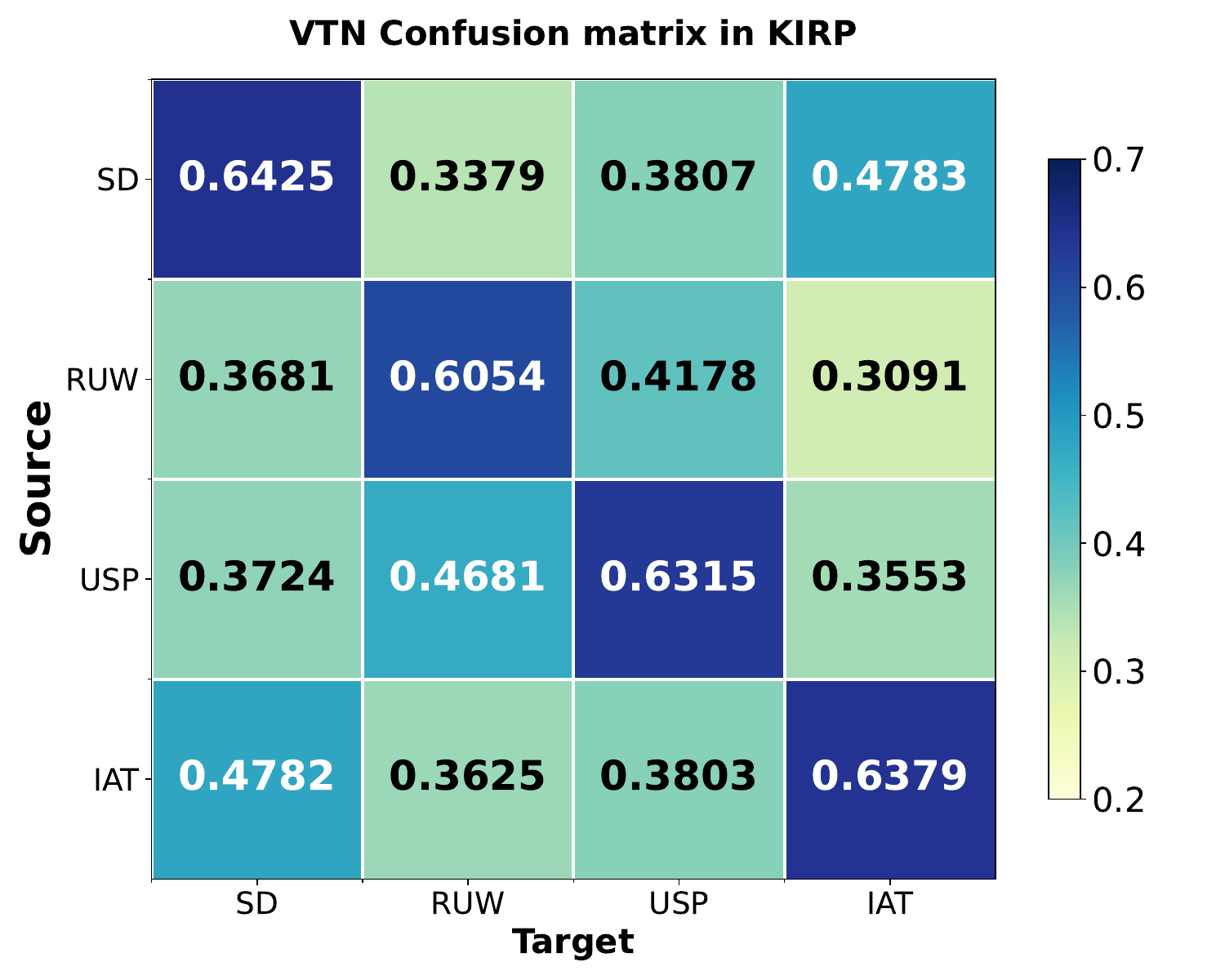}
			\caption*{(c)}
		\end{minipage}
		
		\vspace{0.5em}
		
		\begin{minipage}[c]{0.328\textwidth}
			\centering
			\includegraphics[width=\linewidth]{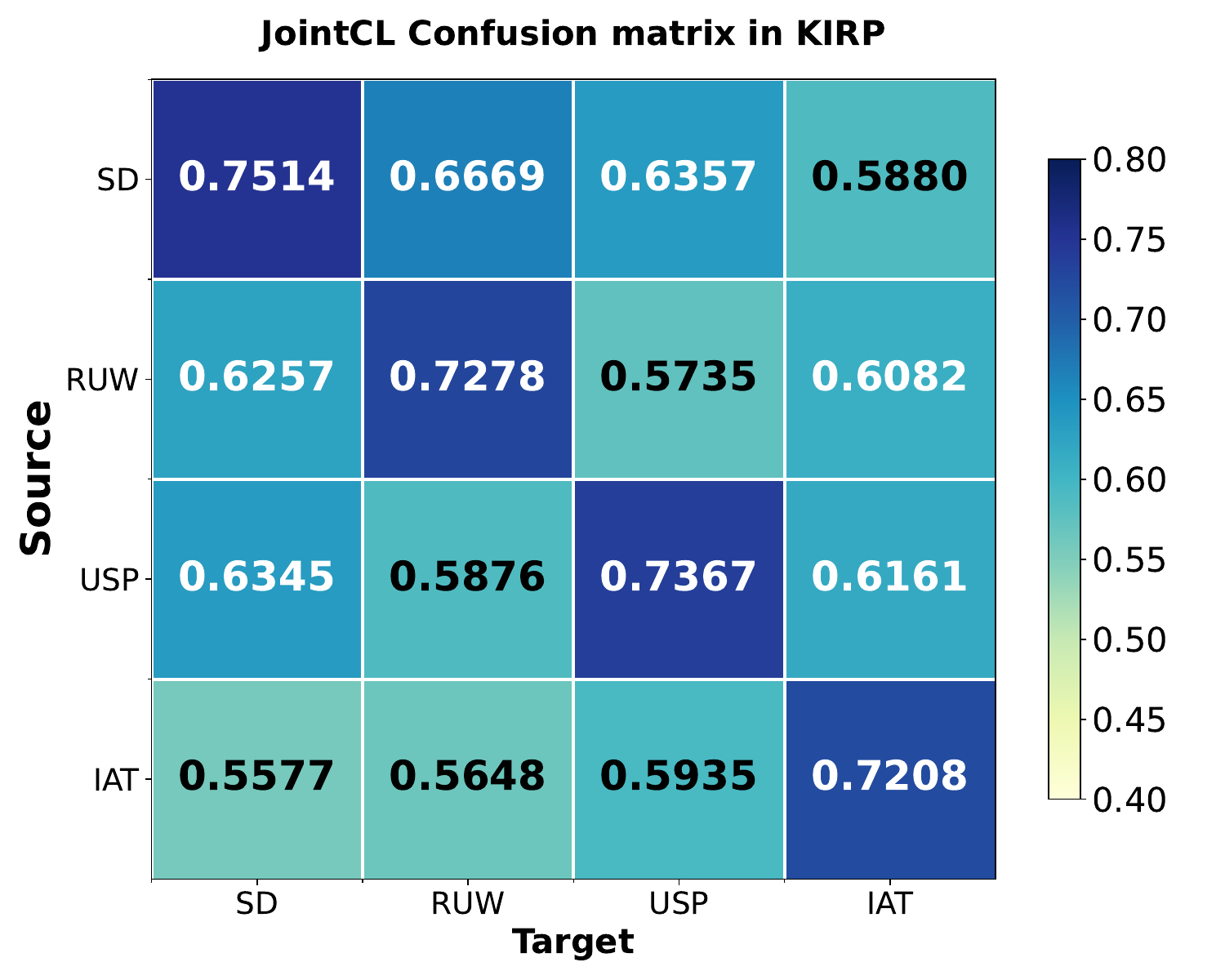}
			\caption*{(d)}
		\end{minipage}
		\hfill
		\begin{minipage}[c]{0.328\textwidth}
			\centering
			\includegraphics[width=\linewidth]{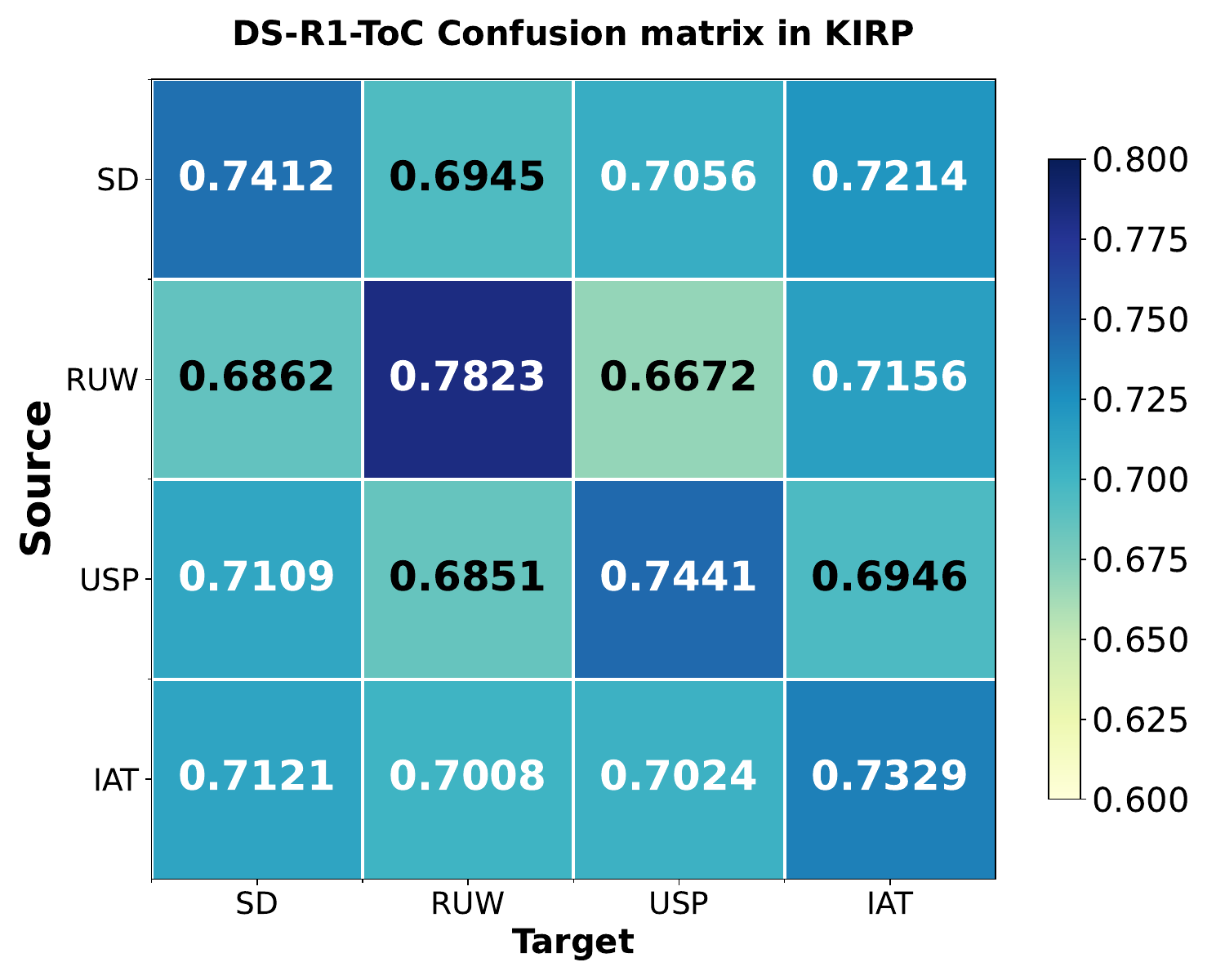}
			\caption*{(e)}
		\end{minipage}
		\hfill
		\begin{minipage}[c]{0.328\textwidth}
			\centering
			\includegraphics[width=\linewidth]{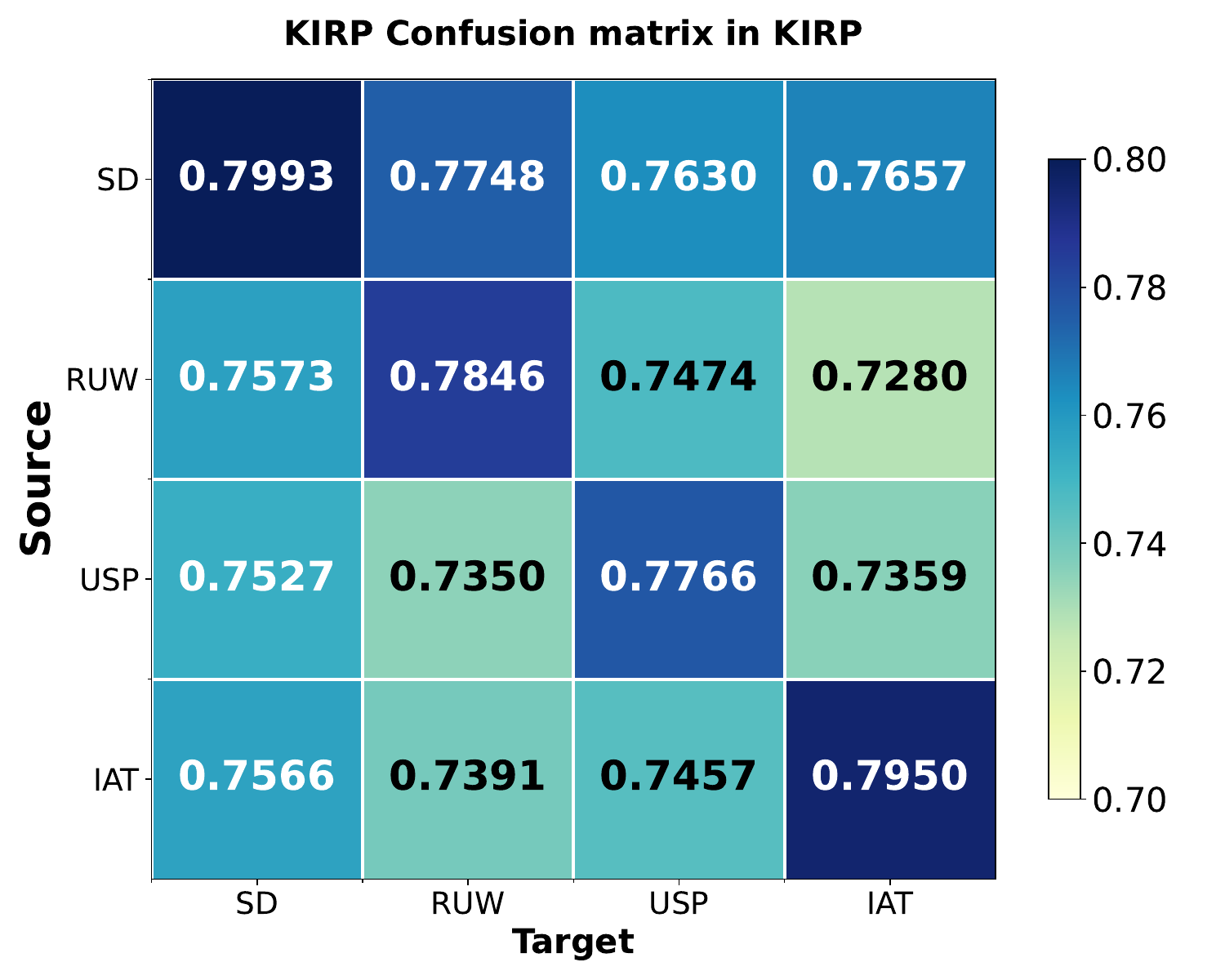}
			\caption*{(f)}
		\end{minipage}
		
		\caption{Comparison of Transfer Performance Across Models on the KIRP Dataset}
		\label{fig:transfer}
	\end{figure*}
	
	To verify that our model possesses cross-target transfer capability, we used the four topics contained in the KIRP dataset (Sewage Discharge, SD; Russia-Ukraine War, RUW; United States Presidential Election, USP; Impose Additional Tariffs, IAT) as known topics and transferred them to three unseen topic scenarios for testing. Representative models from each approach (SKET, BERTweet, VTN, JointCL, DS-R1-ToC, KIRP) were selected for the experiments. The results are shown in the figure below. In the heatmap representation, the vertical axis represents known topics/targets, the horizontal axis represents the unseen topics/targets to which transfer is performed, and the values in the cells are F1 scores. The numerical range on the right side of the heatmap is set to twice the difference between the highest and lowest values to facilitate clarifying the performance of different models. In the heatmap, the closer the colors are to the central symmetry line, the stronger the model's transfer capability; conversely, the more dispersed the colors, the weaker the transfer capability. Results are shown in Fig .5.
	
	According to the experimental results, KIRP exhibits a maximum F1 drop of only 6\% during cross-topic transfer, significantly outperforming other baseline models (DS-R1-ToC at 7.5\%, JointCL at 10\%, VTN at 30\%, BERTweet at 35\%, and SKET at 30\%). Particularly in transfer tasks with large semantic, such as RUW→USP and IAT→SD, KIRP maintains F1 scores above 0.74, with only a 2\% decrease compared to the source training target. DS-R1-ToC, also a large language model prompting chain model, shows relatively small performance degradation as well, at only 3.8\% on average. The experiments confirm that its robustness to topic drift stems from the ToC prompting chain's ability to stably model domain-independent logical structures.
	
	\subsection{Prototypical Network Classification Performance Experiment}
	This experiment investigates the output effectiveness of the prototypical network versus the BERT pre-trained model classification layer for four-class stance classification, comparing the classification performance of the two approaches through embedding visualization. At the top of BERT, there is typically an additional fully connected layer (i.e., classification head) that transforms high-dimensional outputs into forms suitable for specific tasks, and converts classification vectors into probability distributions through the Softmax function, where the probability of each category reflects the model's confidence that the input text belongs to that category. In contrast, the iterative prototypical network transforms this problem into three separate steps, training different prototype spaces to achieve classification.
	
	This experiment targets the iterative prototypical network and the BERT pre-trained model classification layer by removing their output modules, converting them to output only low-dimensional vectors representing classification, and mapping these vectors to two-dimensional space, presenting the classification effects of different classification methods through scatter plots.The performance evaluation compared the classification outcomes of three models—KIRP, PT-HCL, and ANEK—using prototypical networks and the classification layers of BERT-based pretrained models. These are denoted as “-PN” and “-SM,” respectively, in Fig. 6.
	
	As can be seen from the experimental results, the Softmax classification layer can accomplish the classification of “support,” “oppose,” and “neutral” stances, but its discrimination of the “irrelevant” label is poor. The reason is that the “irrelevant” label differs from others only in topic relevance, which cannot be distinguished using stance polarity because it may or may not exist. As shown in Fig. 6, the vectors representing “irrelevant” overlap with those of the other three labels.
	
	To address this issue, this paper introduces a three-layer iterative prototypical network that separates different labels layer by layer to achieve four-class stance classification. First, the FAN/I prototypical network distinguishes topic-irrelevant data points, then the FA/N prototypical network distinguishes stance-neutral points, and finally the F/A prototypical network discriminates stance polarity. The advantage of using the prototypical network lies in its ability to leverage the topic-related characteristics of “support,” “oppose,” and “neutral” data to enhance the representation of topic information, thereby distinguishing topic-irrelevant examples and simplifying the problem to three-class stance classification. In contrast, the BERT classification layer balances topic information with other information (such as knowledge graph information, stance polarity information, etc.), making it difficult to distinguish “irrelevant” stances in the end. This is precisely the limitation of the Softmax classification layer.
	
	\begin{figure*}[!t]
		\centering
		
		\begin{minipage}[c]{0.328\textwidth}
			\centering
			\includegraphics[width=\linewidth]{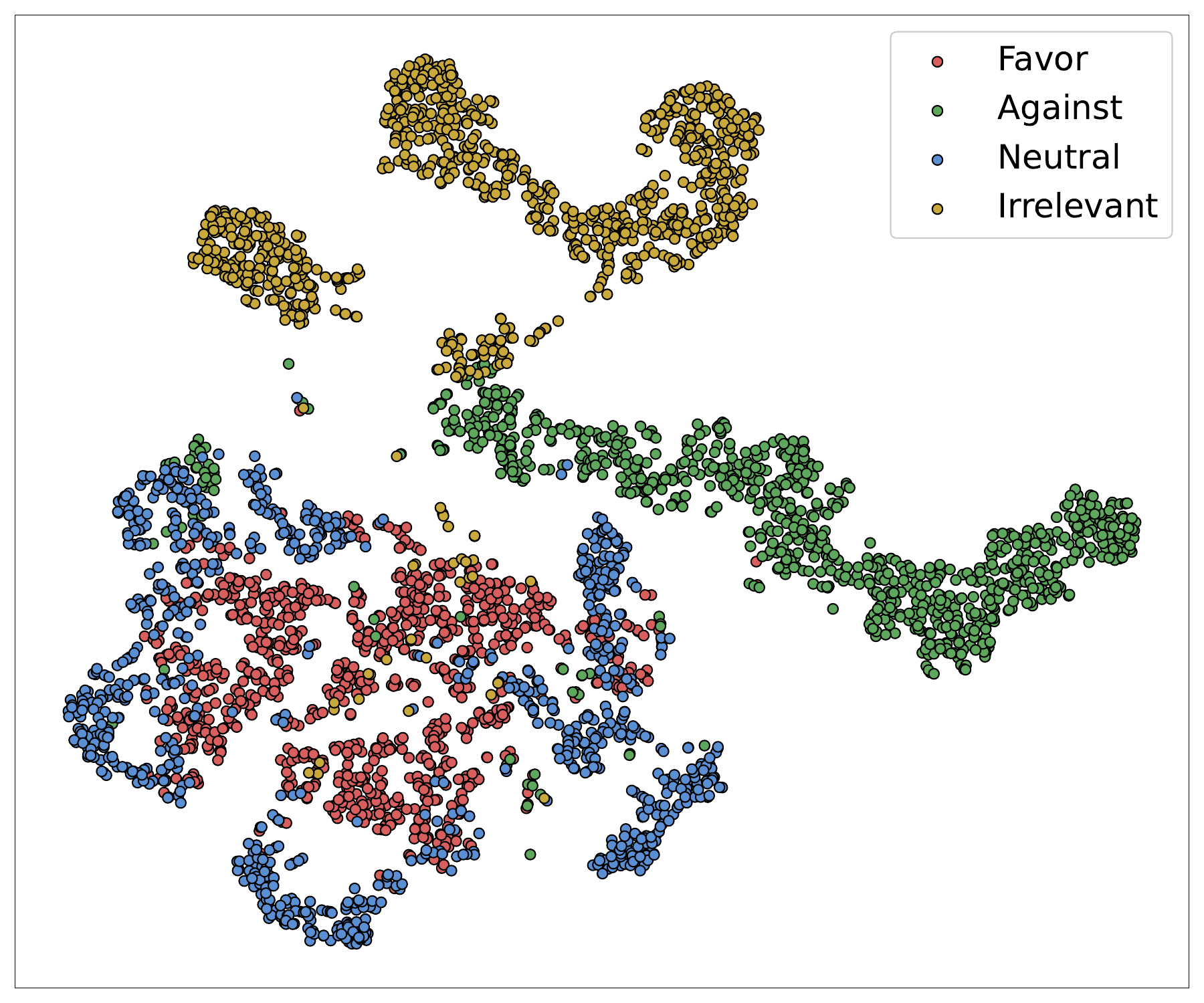}
			\caption*{(a1) KIRP-SM}
		\end{minipage}
		\hfill
		\begin{minipage}[c]{0.328\textwidth}
			\centering
			\includegraphics[width=\linewidth]{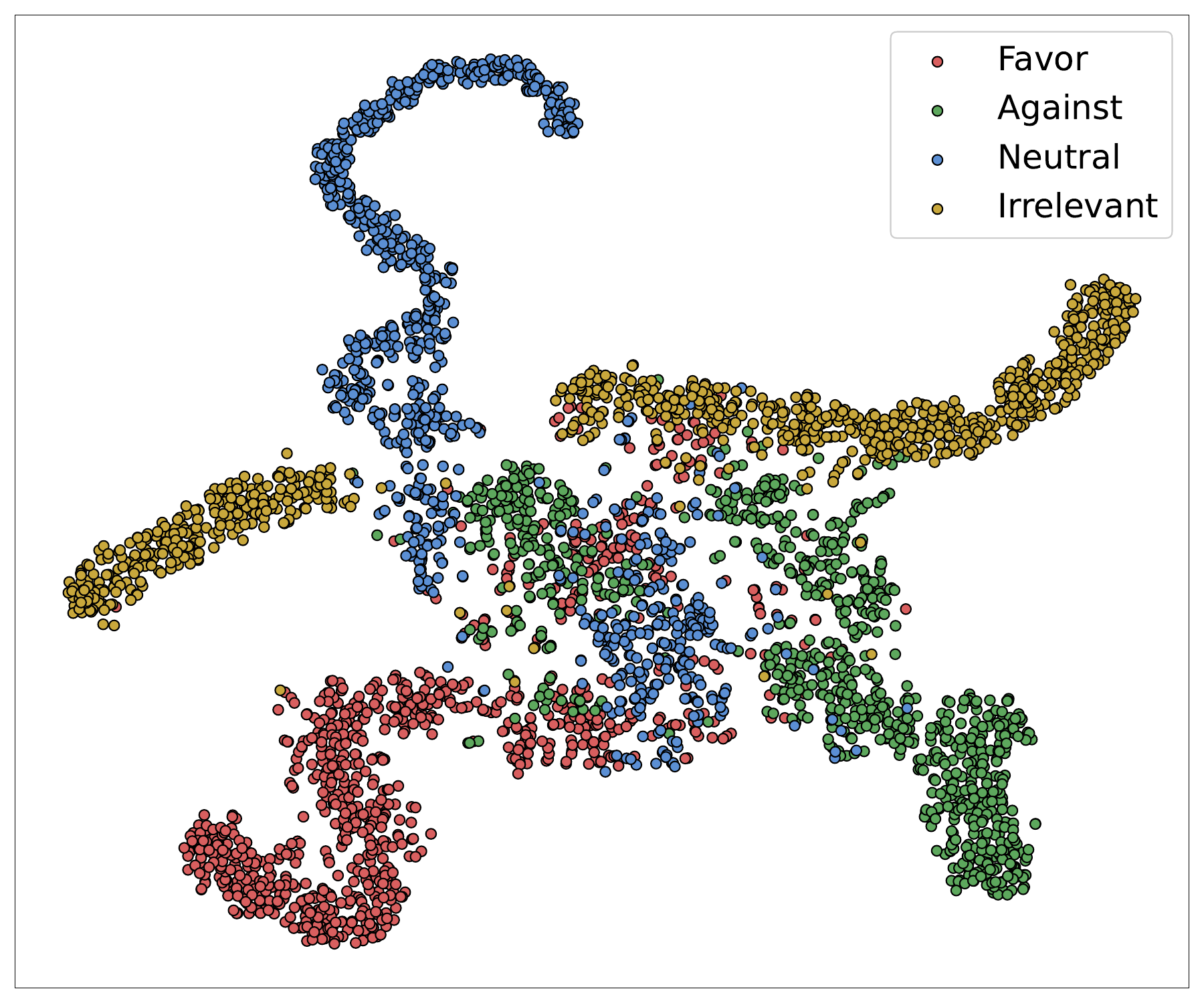}
			\caption*{(b1) PT-HCL-SM}
		\end{minipage}
		\hfill
		\begin{minipage}[c]{0.328\textwidth}
			\centering
			\includegraphics[width=\linewidth]{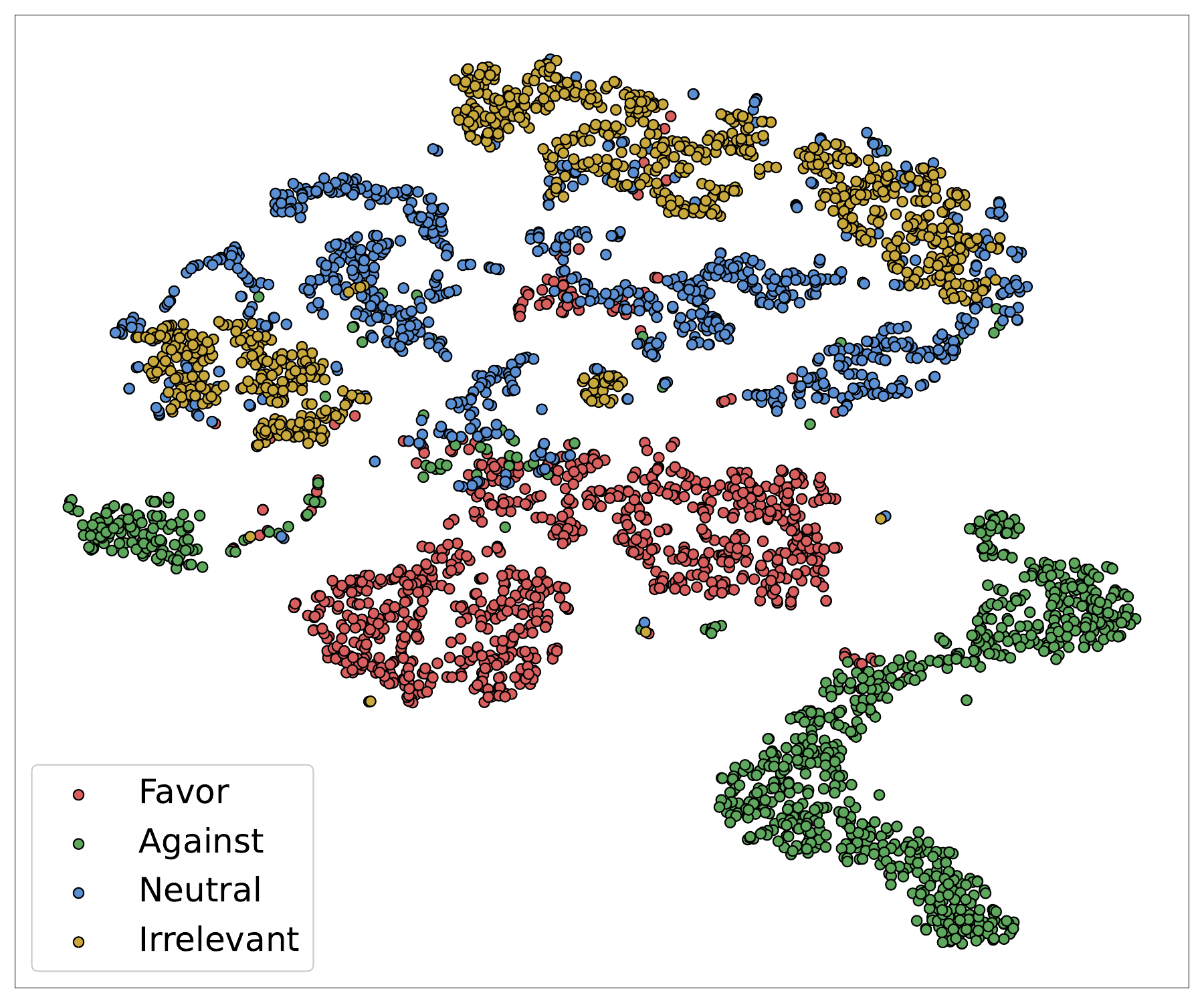}
			\caption*{(c1) ANEK-SM}
		\end{minipage}
		
		\vspace{0.5em}
		
		\begin{minipage}[c]{0.328\textwidth}
			\centering
			\includegraphics[width=\linewidth]{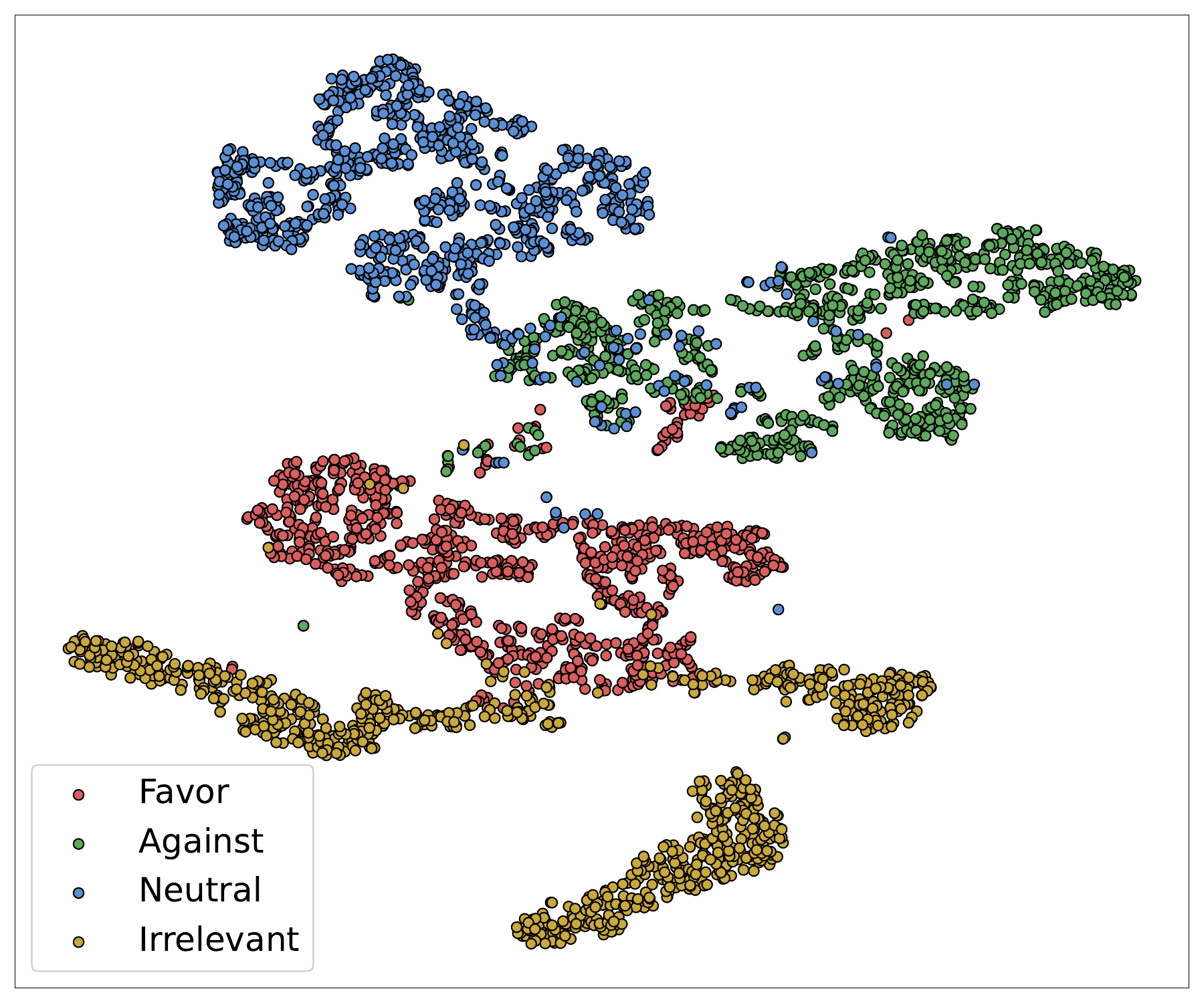}
			\caption*{(a2) KIRP-PN}
		\end{minipage}
		\hfill
		\begin{minipage}[c]{0.328\textwidth}
			\centering
			\includegraphics[width=\linewidth]{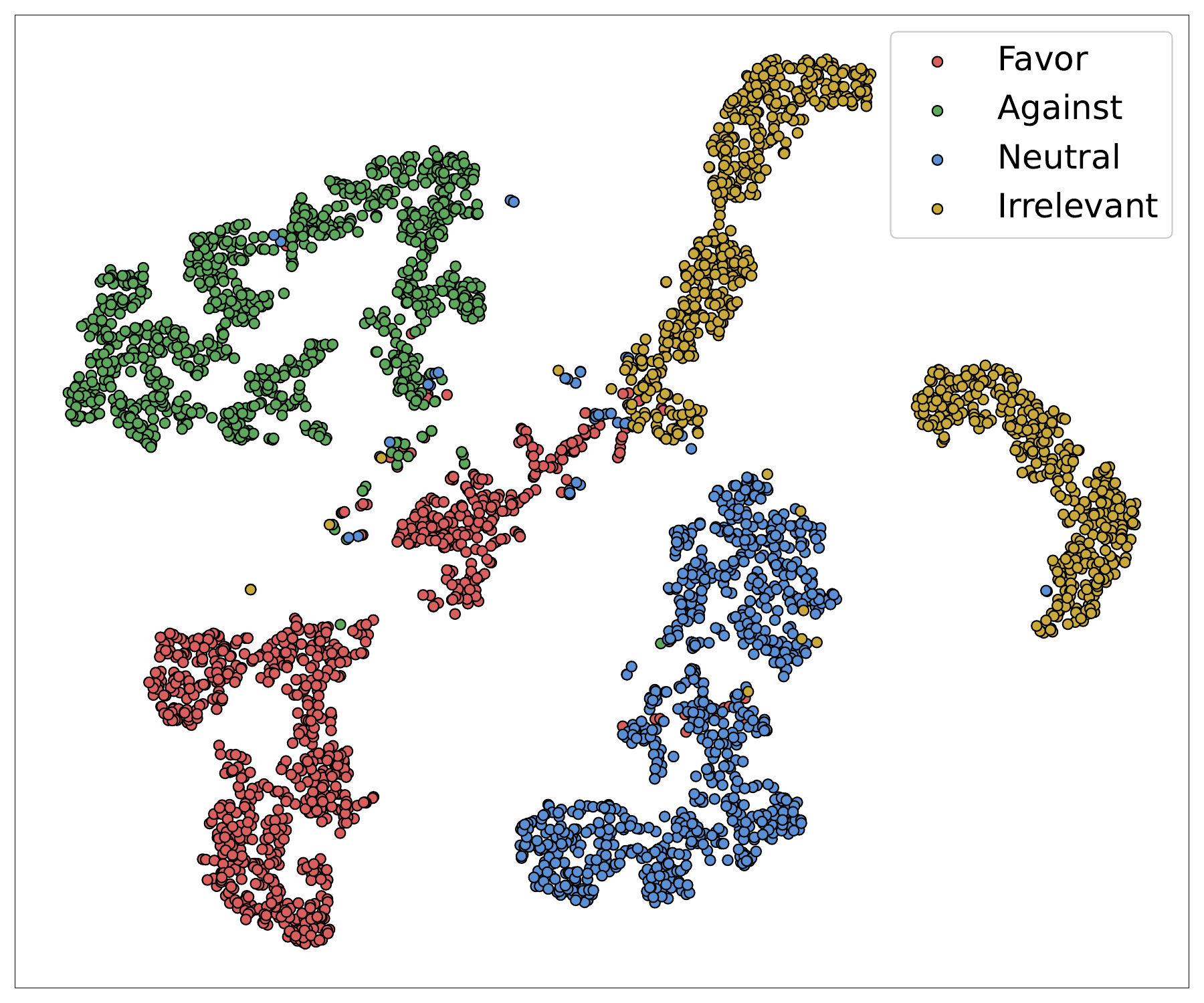}
			\caption*{(b2) PT-HCL-PN}
		\end{minipage}
		\hfill
		\begin{minipage}[c]{0.328\textwidth}
			\centering
			\includegraphics[width=\linewidth]{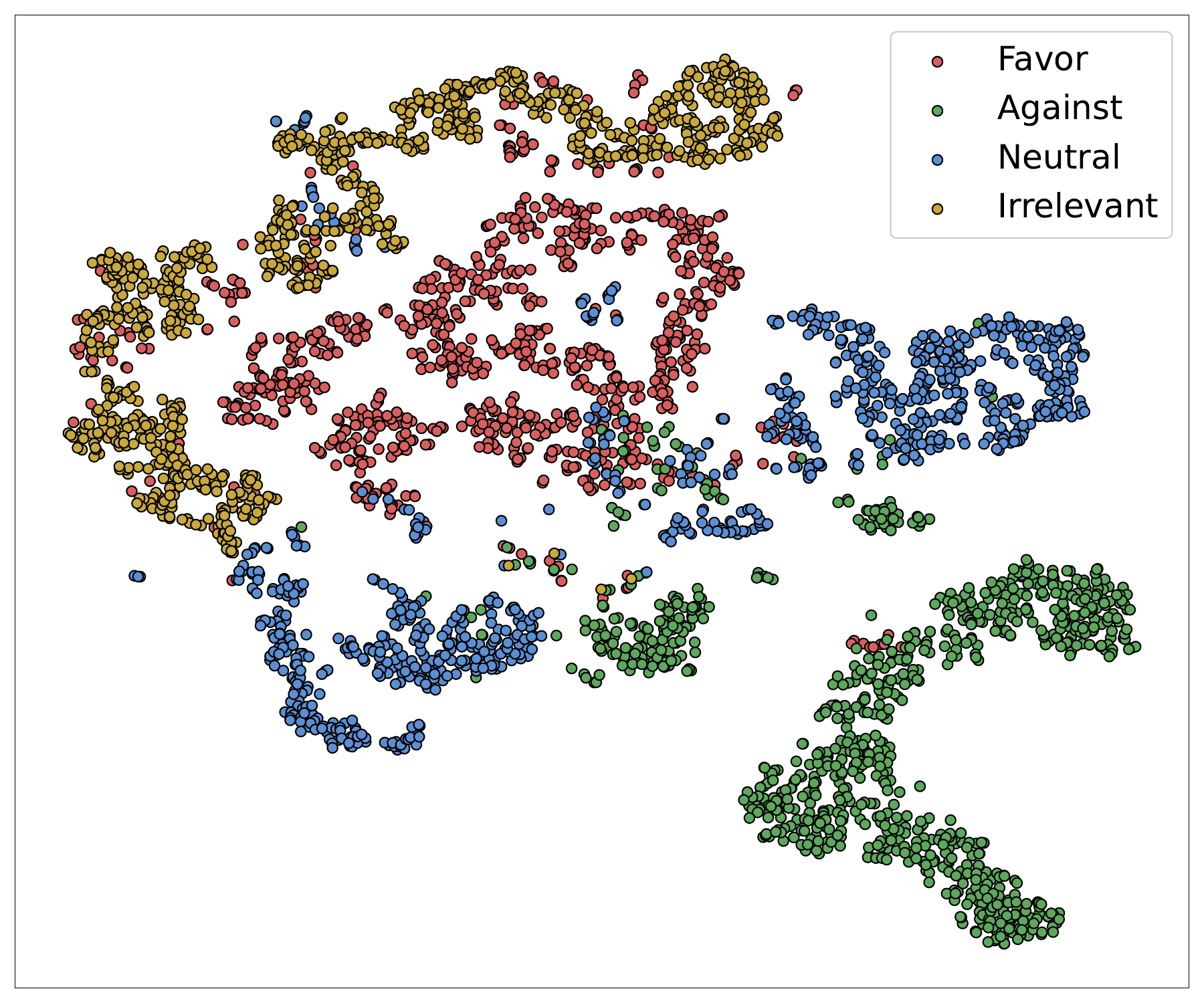}
			\caption*{(c2) ANEK-PN}
		\end{minipage}
		
		\caption{Comparison of Classification Performance between Prototypical Networks and Softmax across Various Models}
		\label{fig:pn_comparison}
	\end{figure*}
	
	\subsection{Case Study}
	During model training, we identified some typical examples. We extracted feedback information from the large language model prompts and collected two representative samples for demonstration, analyzing the logic of the prompting chain classification.
	
	\begin{table*}[h]
		\centering
		\small
		\caption{Appendix 1: Chain-of-Thought Prompting Pipeline}
		\label{tab:case-study}
		\begin{tblr}{
				colspec  = {X[1.2,l,m] X[2.2,l,m] X[3.0,l,m]},
				width    = \textwidth,
				rowsep   = 4pt,
				hline{1} = {-}{0.8pt},
				hline{2} = {-}{0.4pt},
				hline{3} = {-}{0.4pt},
				hline{4} = {-}{0.4pt},
				hline{5} = {-}{0.8pt},
				cell{2}{1} = {r=3}{l,m},
				row{1}   = {font=\bfseries\small, halign=c},
				vline{2,3} = {-}{0.4pt},
			}
			Text & User (Prompt) & LLM (Response) \\
			{
				\textbf{Text:}\\
				\textit{Industrial waste liquid, fuel debris\ldots\ These wonderful elements will be unhesitatingly poured into rivers and seas by our government. Isn't this a form of energy conservation? We can save a significant sum of money for officials by avoiding the need to hire specialized cleaning personnel, while also bringing citizens a mother river as nurturing as the Ganges. My fish friends in the river and I both support this selfless practice---I will soon be immigrating to New Jersey, but I will miss my hometown friends.}\\[4pt]
				\textbf{Target:}\\
				\textit{sewage discharge}\\[4pt]
				\textbf{Stance:}\\
				\textit{Favor}
			}
			&
			{
				\textbf{Stage 1: Generate Criteria}\\[3pt]
				\textit{You are an expert in the field of stance detection, and there is a new type of stance detection work that requires your cooperation. What you need to do is: Generate a possible reasonable explanation for the stance of Text} \underline{\textit{[Text]}} \textit{being} \underline{\textit{[Stance]}}\textit{, without considering whether the stance is reasonable or not.}\\[2pt]
				\textit{You need to generate a reply in the following format: The following are three possible reasonable explanations for Text [text]'s stance on Target [target] being [stance].}\\[2pt]
				\textit{1. explanation 1}\\
				\textit{2. explanation 2}\\
				\textit{3. explanation 3\ldots}
			}
			&
			{
				\textit{The following are three possible reasonable explanations for Text} \underline{\textit{[Text]}}\textit{'s stance on Target} \underline{\textit{[Target]}} \textit{being} \underline{\textit{[Stance]}}\textit{.}\\[3pt]
				\textit{1.}~\textbf{\textit{Irony and Satire:}} \textit{The speaker might be using heavy irony to exaggerate the government's actions, possibly criticizing sewage discharge while pretending to endorse it. The explanation suggests that by using the word ``wonderful'' and equating it to a supposed ``form of energy conservation,'' the speaker may be mocking the government's irresponsible practices while pretending to show support.}\\[2pt]
				\textit{2.}~\textbf{\textit{Absurdist Humor:}} \textit{The speaker might be employing absurdist humor, portraying the sewage discharge as a bizarre, yet beneficial practice. The reference to ``saving money for officials'' and the exaggerated comparison to the Ganges river could indicate that the speaker is playfully supporting the discharge to ridicule the situation.}\\[2pt]
				\textit{3.}~\textbf{\textit{Satirical Support:}} \textit{The speaker's stance might be interpreted as favoring sewage discharge through a satirical lens, where they claim that the government's actions are a form of ``energy conservation'' and financially advantageous.}
			}
			\\
			&
			{
				\textbf{Stage 2: Reflection}\\[3pt]
				\textit{Based on the above explanation, please reflect on whether the position of text} \underline{\textit{[text]}} \textit{towards target} \underline{\textit{[target]}}\textit{, which is} \underline{\textit{[stance]}}\textit{, is reasonable. If not, please point out the un-reasonable aspects of the above explanation.}\\[2pt]
				\textit{Also you need to generate a reply in the following format: The stance of text [text] towards target [target] is that [stance] is unreasonable, and the above explanation has the following un-reasonable aspects}\\[2pt]
				\textit{1. reflection 1}\\
				\textit{2. reflection 2}\\
				\textit{3. reflection 3}
			}
			&
			{
				\textit{The stance of text} \underline{\textit{[text]}} \textit{towards target} \underline{\textit{[target]}} \textit{being} \underline{\textit{[stance]}} \textit{is \textbf{unreasonable}, and the above explanation has the following unreasonable aspects:}\\[3pt]
				\textbf{\textit{1. Contradictory Support:}} \textit{The explanation seems to ignore the contradiction within the text itself. While the speaker claims to support sewage discharge, they use phrases like ``wonderful elements'' and ``mother river as nurturing as the Ganges'' in a clearly exaggerated, absurd manner. These statements seem to mock, rather than truly endorse, the practice of sewage discharge.}\\[2pt]
				\textbf{\textit{2. Failure to Acknowledge Criticism:}} \textit{The explanation fails to account for the critical undertone of the text. By emphasizing ``saving a significant sum of money for officials'' and ``avoiding the need to hire specialized cleaning personnel,'' the speaker draws attention to the neglect and irresponsibility of the government's actions.}\\[2pt]
				\textbf{\textit{3. Inconsistent Use of Humor and Sarcasm:}} \textit{The explanation does not fully recognize that the text likely uses absurdist humor or sarcasm to mock the sewage discharge. The reference to ``fish friends'' and the ridiculous comparison to the Ganges river highlight an ironic stance, meant to expose the absurdity of the situation.}
			}
			\\
			&
			{
				\textbf{Stage 3: Correct}\\[3pt]
				\textit{How to modify the content of text [text], target [target], and stance [favor] in order to establish stance [favor]. You can only modify some nouns in the text. Please use \underline{underline} or \textbf{bold} to distinguish the modified vocabulary from the original text.}\\[2pt]
				\textit{Also you need to generate a reply in the following format: The possible modifications are as follows:}\\[2pt]
				\textit{1. [text1]$\rightarrow$[text2]}\\
				\textit{2. [target1]$\rightarrow$[target2]}\\
				\textit{3. [stance1]$\rightarrow$[stance2]}
			}
			&
			{
				\textit{The possible modifications are as follows:}\\[3pt]
				\textit{[text1] $\rightarrow$ ``Industrial waste liquid, fuel debris\ldots\ These \textbf{important} elements will be \textbf{carefully managed} by our government. Isn't this a form of \textbf{environmental protection}? We can save a significant sum of money for officials by avoiding the need to hire specialized cleaning personnel, while also bringing citizens a \textbf{clean and healthy} river, providing a \textbf{safe} environment for all living creatures. My \textbf{community} and I both support this \textbf{effective} practice---this is the right way forward for our future.''}\\[3pt]
				\textit{[target1] $\rightarrow$ \textbf{[environmental management]}}\\[3pt]
				\textit{[stance1] $\rightarrow$ \textbf{[favor]}}
			}
			\\
		\end{tblr}
	\end{table*}
	
	In Case Study 1, the given data contains an example with ironic statements, and the hypothetical stance value assigned is incorrect. In the criteria supplementation stage, the model provides a criterion hypothesis of “irony and sarcasm,” suggesting that expressions in the text that deviate from conventional logic may imply sarcastic meaning. This is because environmental protection is widely regarded as just, leading the large language model to assume support for it rather than random sewage discharge. Therefore, the model quickly recognizes the ironic undertone in this passage.
	
	In the second stage of the reflection process, the model provides three reflective arguments to support its viewpoint. Finally, in the correction stage, the ironic statements in the original text are removed. Based on the observation that the large language model tends to make judgments based on socially common-sense conclusions, we conducted a second case experiment, as shown in the Table IX.
	
	In Case Study 2, the given data also contains an example with ironic statements, and the hypothetical stance value assigned is incorrect. In the criteria supplementation stage, the model interprets the text as praising Trump's economic capabilities and supplements corresponding criterion information. However, during the reflection stage, the model identifies the issue and provides prompts indicating that the stance label is unreasonable. The model considers that the description in this passage over-glorifies business acumen, based on the assumption that Trump's business background would automatically translate into political capability. In reality, the logical connection between business background and political capability is not that strong, and the expression in this passage overlooks the complexity of governing a country.
	
	The large language model utilizes common-sense knowledge—“not all successful business leaders can become successful politicians”—to analyze whether the excessive praise of Trump's business abilities in the text is actually sarcasm about his inadequate political capabilities.
	
	In the third reflective argument, the model begins to consider the text's stance toward the implicit target—Harris. Based on the text's expression, the model considers that “Harris lacks business experience and therefore is unqualified to become a business president” reflects a biased stance based on a single characteristic, underestimating her potential. Based on this reasoning, the model questions whether the text might tend to support Harris, as it does not mention Harris's political capabilities.
	
	\begin{table*}[h]
		\centering
		\small
		\caption{Appendix 2: Chain-of-Thought Prompting Pipeline}
		\label{tab:case-study-2}
		\begin{tblr}{
				colspec  = {X[1.2,l,m] X[2.2,l,m] X[3.0,l,m]},
				width    = \textwidth,
				rowsep   = 4pt,
				hline{1} = {-}{0.8pt},
				hline{2} = {-}{0.4pt},
				hline{3} = {-}{0.4pt},
				hline{4} = {-}{0.4pt},
				hline{5} = {-}{0.8pt},
				cell{2}{1} = {r=3}{l,m},
				row{1}   = {font=\bfseries\small, halign=c},
				vline{2,3} = {-}{0.4pt},
			}
			Text & User (Prompt) & LLM (Response) \\
			{
				\textbf{Text:}\\
				\textit{I believe Trump will win this election because there is no official in the US government who understands business as well as he does. They are all waiting for Trump to bring them endless wealth. Perhaps Harris will become the first female president in American history, but she will never become the first business president in American history, or a presidential businessman.}\\[4pt]
				\textbf{Target:}\\
				\textit{Trump}\\[4pt]
				\textbf{Stance:}\\
				\textit{Favor}
			}
			&
			{
				\textbf{Stage 1: Generate Criteria}\\[3pt]
				\textit{You are an expert in the field of stance detection, and there is a new type of stance detection work that requires your cooperation. What you need to do is:}\\[2pt]
				\textit{Generate a possible reasonable explanation for the stance of Text} \underline{\textit{[Text]}} \textit{towards the target} \underline{\textit{[Trump]}} \textit{being} \underline{\textit{[favor]}}\textit{, without considering whether the stance is reasonable or not.}\\[2pt]
				\textit{You need to generate a reply in the following format: The following are three possible reasonable explanations for Text [text]'s stance on Target [target] being [stance].}\\[2pt]
				\textit{1. explanation 1}\\
				\textit{2. explanation 2}\\
				\textit{3. explanation 3}
			}
			&
			{
				\textit{The following are three possible reasonable explanations for Text} \underline{\textit{[Text]}}\textit{'s stance on Target} \underline{\textit{[Trump]}} \textit{being} \underline{\textit{[favor]}}\textit{.}\\[3pt]
				\textit{1. The text emphasizes Trump's unique expertise in business, contrasting it with the perceived lack of similar qualifications among other politicians, positioning him as the only candidate capable of driving wealth creation and economic prosperity, which might inspire voters to favor him.}\\[2pt]
				\textit{2. The statement suggests that Trump's business acumen is the key to his success, with the implication that his presidency would bring material benefits to both individuals and the country, making him the preferred candidate for those seeking business-driven governance.}\\[2pt]
				\textit{3. By noting Harris' gender and suggesting that she lacks Trump's business credentials, the text underscores Trump's distinctive identity as a ``business president.'' This creates a clear favor for Trump, positioning him as a candidate with a unique and valuable skill set that others, including Harris, do not possess.}
			}
			\\
			&
			{
				\textbf{Stage 2: Reflection}\\[3pt]
				\textit{Based on the above explanation, please reflect on whether the position of text} \underline{\textit{[text]}} \textit{towards target} \underline{\textit{[target]}}\textit{, which is} \underline{\textit{[stance]}}\textit{, is reasonable. If not, please point out the un-reasonable aspects of the above explanation.}\\[2pt]
				\textit{Also you need to generate a reply in the following format: The stance of text [text] towards target [target] is that [stance] is unreasonable, and the above explanation has the following un-reasonable aspects}\\[2pt]
				\textit{1. reflection 1}\\
				\textit{2. reflection 2}\\
				\textit{3. reflection 3}
			}
			&
			{
				\textit{The stance of text} \underline{\textit{[Text]}}\textit{'s stance towards target} \underline{\textit{[Trump]}} \textit{being} \underline{\textit{[favor]}} \textit{is \textbf{unreasonable}, and the above explanation has the following unreasonable aspects:}\\[3pt]
				\textbf{\textit{1. Overgeneralization of business acumen:}} \textit{The text assumes that Trump's business background automatically translates to political success, overlooking the complexity of governance. Not all effective leaders in business make successful politicians, and this simplistic view disregards other qualities that contribute to good leadership, such as diplomacy, social policy understanding, and international relations.}\\[2pt]
				\textbf{\textit{2. Disregard for other qualifications:}} \textit{By focusing solely on business acumen as the determinant for success, the text undermines other critical factors necessary for running a country. The exclusion of important political experience or the ability to manage diverse policy issues makes the stance unreasonable.}\\[2pt]
				\textbf{\textit{3. Bias and negative comparison:}} \textit{The text suggests that Harris' lack of business experience disqualifies her from becoming a ``business president.'' This is a biased stance that discounts her potential based on a single characteristic. It also dismisses other aspects of leadership, such as policy expertise, social justice work, and the ability to unite a country, which are all crucial for presidential success.}
			}
			\\
			&
			{
				\textbf{Stage 3: Correct}\\[3pt]
				\textit{How to modify the content of text [text], target [target], and stance [favor] in order to establish stance [favor]. You can only modify some nouns in the text. Please use \underline{underline} or \textbf{bold} to distinguish the modified vocabulary from the original text.}\\[2pt]
				\textit{Also you need to generate a reply in the following format: The possible modifications are as follows:}\\[2pt]
				\textit{1. [text1]$\rightarrow$[text2]}\\
				\textit{2. [target1]$\rightarrow$[target2]}\\
				\textit{3. [stance1]$\rightarrow$[stance2]}
			}
			&
			{
				\textit{The possible modifications are as follows:}\\[3pt]
				\textit{[text1] $\rightarrow$ [I believe Trump will win this election because there is no official in the US government who understands \textbf{economics} as well as he does. They are all waiting for Trump to bring them endless \textbf{prosperity}. Perhaps Harris will become the first female president in American history, but she will never become the first \textbf{economic} president in American history, or a \textbf{business-oriented} president.]}\\[3pt]
				\textit{[Trump] $\rightarrow$ [Trump]}\\[3pt]
				\textit{[favor] $\rightarrow$ [favor]}
			}
			\\
		\end{tblr}
	\end{table*}
	
	In the third stage of correction, the large language model synthesizes the arguments from the previous two stages but ultimately concludes that the text's stance is support for Trump. In reality, the stance of this text is “neutral.” The reason for the model's misjudgment lies in its over-reliance on common-sense information, assuming that Trump and Harris are completely opposing figures, while overlooking the possibility that someone might support neither of them. The model successfully identified the sarcasm toward Trump in the text, thus determining that the text only superficially supports Trump. However, in actuality, the author of this text does not support either of them, as the text both expresses sarcasm toward Trump and does not express support for Harris winning the election. Therefore, the stance of this text should be “neutral.”
	
	\section{Limitations}
	Although KIRP achieves state-of-the-art performance across multiple datasets, certain limitations remain. From a data perspective, KIRP-D currently covers four topics across Japanese social media, and its generalizability to a broader range of languages and domains may warrant further investigation. From a model perspective, the reflective CoT reasoning module depends on the capacity of the underlying LLM, meaning that variations in LLM behavior across different model versions or domains could influence the consistency of generated reasoning criteria. Additionally, the external knowledge graph is constructed from Wikipedia and news sources, whose coverage and timeliness may vary across topics, potentially affecting the quality of entity augmentation in highly specialized or rapidly evolving scenarios. These aspects present opportunities for future exploration.
	
	\section{Conclusion}
	This paper proposes an external knowledge-enhanced zero-shot tweet-level stance detection model and constructs the first Japanese stance detection dataset for this task by collecting data from the Twitter platform. Specifically, a graph autoencoder is employed to embed the knowledge graph, while a reflective CoT is introduced to generate stance criteria and perform preliminary classification. Meanwhile, stance contrastive learning is utilized to enhance textual representations. By combining these components with an iterative prototype network-based binary classification strategy, the proposed model addresses the four-class stance detection task. Experimental results demonstrate the effectiveness of the proposed approach. Comparative and ablation studies validate its superiority, and case studies are conducted to analyze and explain representative error cases.
	
	\normalem
	\bibliographystyle{IEEEtran}
	\bibliography{Bibsample}
	
\end{document}